\documentclass[a4paper,fleqn]{cas-sc}

\usepackage[numbers]{natbib}

\newcommand{\ours}{HERALD}

\begin{document}
\let\WriteBookmarks\relax
\def\floatpagepagefraction{1}
\def\textpagefraction{.001}

\shorttitle{Forecasting the Emergence and Evolution of Crash Hotspots}

\shortauthors{J. Zhu et~al.}

\title[mode = title]{Forecasting the Emergence and Evolution of Crash Hotspots: A Unified Deep Learning Framework for Proactive Traffic Safety}

\author[1]{Jingwen Zhu}
\ead{jzhu254@wisc.edu}
\credit{Conceptualization, Data curation, Formal analysis, Writing -- original draft}

\author[2,1]{Keshu Wu}
\ead{keshuw@tamu.edu}
\credit{Conceptualization, Formal analysis, Writing -- review \& editing}

\author[3,1]{Pei Li}
\ead{pei.li@wisc.edu}
\credit{Formal analysis, Writing -- review \& editing}

\author[1]{Steven T. Parker}
\ead{sparker@engr.wisc.edu}
\credit{Conceptualization, Data curation, Formal analysis, Writing -- review \& editing}

\author[1]{Bin Ran}
\ead{bran@wisc.edu}
\credit{Supervision}

\author[1]{David A. Noyce}
\ead{noyce@engr.wisc.edu}
\credit{Supervision}


\affiliation[1]{organization={Department of Civil and Environmental Engineering, University of Wisconsin--Madison},
    city={Madison},
    state={WI},
    country={United States}}

\affiliation[2]{organization={Zachry Department of Civil and Environmental Engineering, Texas A\&M University},
    city={College Station},
    state={TX},
    country={United States}}

\affiliation[3]{organization={Department of Civil and Architectural Engineering, University of Wyoming},
    city={Laramie},
    state={WY},
    country={United States}}

\begin{abstract}
Road crashes remain among the gravest threats to public safety, and preventing them is a defining task of transportation systems worldwide. Much of that harm concentrates at hotspots, yet a hotspot is less a fixed place than an evolving episode: it emerges at an intersection or along an arterial, intensifies, persists, and eventually subsides or reappears elsewhere. Enforcement guided only by maps of past crashes can trail this cycle, directing attention to yesterday's hotspots while new risks develop elsewhere. Addressing this lag requires three capabilities: detecting emerging hotspots, forecasting their near-term spatial distribution, and tracking their evolution over time. We introduce HERALD (Hotspot Emergence, Risk Anticipation, and Life-cycle Dynamics), a unified deep learning framework that supports all three within a single statewide model. HERALD represents each county's recent crash history as weekly risk maps and forecasts the next rolling window using a CNN-Transformer with a mixture-of-experts that adapts to dense urban cores and sparse rural corridors. Each forecast is anchored in the county's long-run crash geography, refined through the self-exciting influence of recent crashes, and accompanied by warnings of potential hotspot emergence. A matching procedure links detected clusters across time and summarizes their progression through birth, growth, stability, decline, and death. Across six heterogeneous Wisconsin counties, HERALD achieves lower grid-level forecasting error and more accurate overall hotspot localization than five identically trained baselines. Its calibration provides an explicit trade-off between grid accuracy and detection sensitivity for different deployment priorities. The resulting framework advances hotspot management from retrospective mapping toward proactive, interpretable crash-risk forecasting.
\end{abstract}



\maketitle

\section{Introduction}\label{sec:introduction}

Each year, road-traffic crashes claim nearly 1.2 million lives and injure as many as fifty million people~\cite{world2023global, peden2004world}. In the United States alone, they account for tens of thousands of deaths and hundreds of billions of dollars in healthcare, legal, and productivity losses annually~\cite{blincoe2015economic, Corsaro2012}. Behind each of these statistics is a place: a corridor, an intersection, a stretch of highway where risk built up quietly until it turned into harm. The toll is not inevitable: decades of safety research show that well-targeted countermeasures prevent crashes~\cite{elvik2009handbook}. Road-safety policy has accordingly coalesced around Vision Zero, which regards every traffic death and serious injury as preventable and shifts responsibility from reacting to crashes toward anticipating them~\cite{tingvall1999vision, belin2012vision, kim2017vision}. Anticipation, however, is an operational challenge as much as an ethical stance. It requires knowing where danger is gathering before harm occurs, so that scarce enforcement and engineering resources arrive in time.

What makes such anticipation feasible is that crash risk is highly concentrated: a small share of locations and time windows accounts for a disproportionate fraction of severe crashes~\cite{AbdelAty2004, getis1992analysis}. A location of persistently elevated crash risk, relative to its traffic exposure, is termed a \emph{hotspot}, and identifying and prioritizing hotspots is the traditional basis for targeted enforcement and engineering. Where and when resources are directed against them therefore largely determines how many crashes are prevented. Exploiting this concentration is nonetheless hard, because crashes are rare events~\cite{li2025simulating}. At the resolution needed for targeted intervention, most grid cells and most weeks contain no crash at all, and fatal or incapacitating outcomes are rarer still. The resulting signal is sparse, over-dispersed, and highly imbalanced, and it destabilizes naive predictors~\cite{lord2010statistical, mannering2014analytic, mannering2016unobserved}. A useful system must therefore know not only where risk has been, but where it is forming next and how long it will persist, while staying calibrated under this scarcity rather than hallucinating risk in quiet areas or smoothing away the rare cells where crashes concentrate~\cite{parsa2020toward}.

Current practice does not yet meet this bar, on either the analytical or the operational side. The dominant hotspot tools are retrospective and static. Multi-year kernel density estimation and uniform grid aggregation summarize where crashes have clustered in the past~\cite{AbdelAty2004}, and refinements that adapt kernel density to road networks or fuse it with clustering sharpen these maps~\cite{xie2008kernel, anderson2009kernel}. The resulting rankings, however, shift with the identification method chosen~\cite{cheng2005experimental}, and all of them aggregate over long horizons. They blur transient surges, do not separate emerging clusters from dissipating ones, and leave practitioners little to act on in near real time~\cite{sam2022effective, riyanto2020implementation}.

Operational programs have pushed beyond static maps toward data-driven anticipation. Predictive policing pioneered this shift, forecasting where criminal activity will concentrate and steering patrols toward evolving ``\textit{hotspots}''~\cite{Nyce2007, Weisburd2008, Hardy2010}. Systematic reviews link such hotspot policing to measurable reductions in crime~\cite{Braga2014, NASEM2018}, and the approach continues to evolve toward micro-scale and emerging hotspots~\cite{wu2021everyday, mahmoud2021vulnerable, li2023combining, park2025micro}. Road safety adopted the same logic through the Data-Driven Approaches to Crime and Traffic Safety initiative, which fuses crash and enforcement data to cut both crime and traffic incidents~\cite{NHTSA2014}, and spatio-temporal crash patterning now informs resource allocation across several jurisdictions~\cite{wu2023exploring, kaygisiz2015spatio, bil2019detailed}. In the same tradition, the Community Maps platform, built by the Traffic Operations and Safety Laboratory at the University of Wisconsin--Madison with the Wisconsin State Patrol, couples interactive crash mapping with hotspot detection for high-visibility enforcement~\cite{Burrell2018, WisconsinCommunityMaps2020}, and supports routine deployment today~\cite{Wolfe2018, Williams2020, akuh2023impact}. These platforms nonetheless remain retrospective, seldom separate growing from declining clusters, transfer poorly between urban and rural regimes, and give no explicit account of hotspot life cycles.

\begin{figure}[pos=!t]
  \centering
  \begin{minipage}[b]{0.48\textwidth}
    \centering
    \includegraphics[width=\linewidth]{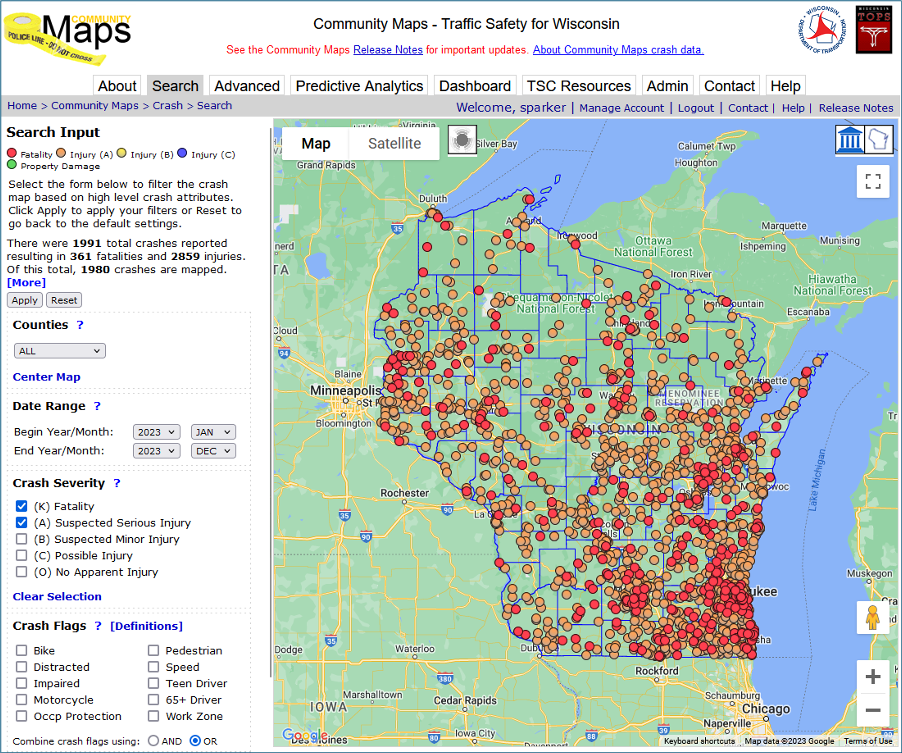}\\
    {\small \textbf{a} Spatial Hotspot Heatmap.}
  \end{minipage}
  \hspace{0.5em}
  \begin{minipage}[b]{0.24\textwidth}
    \centering
    \includegraphics[width=\linewidth]{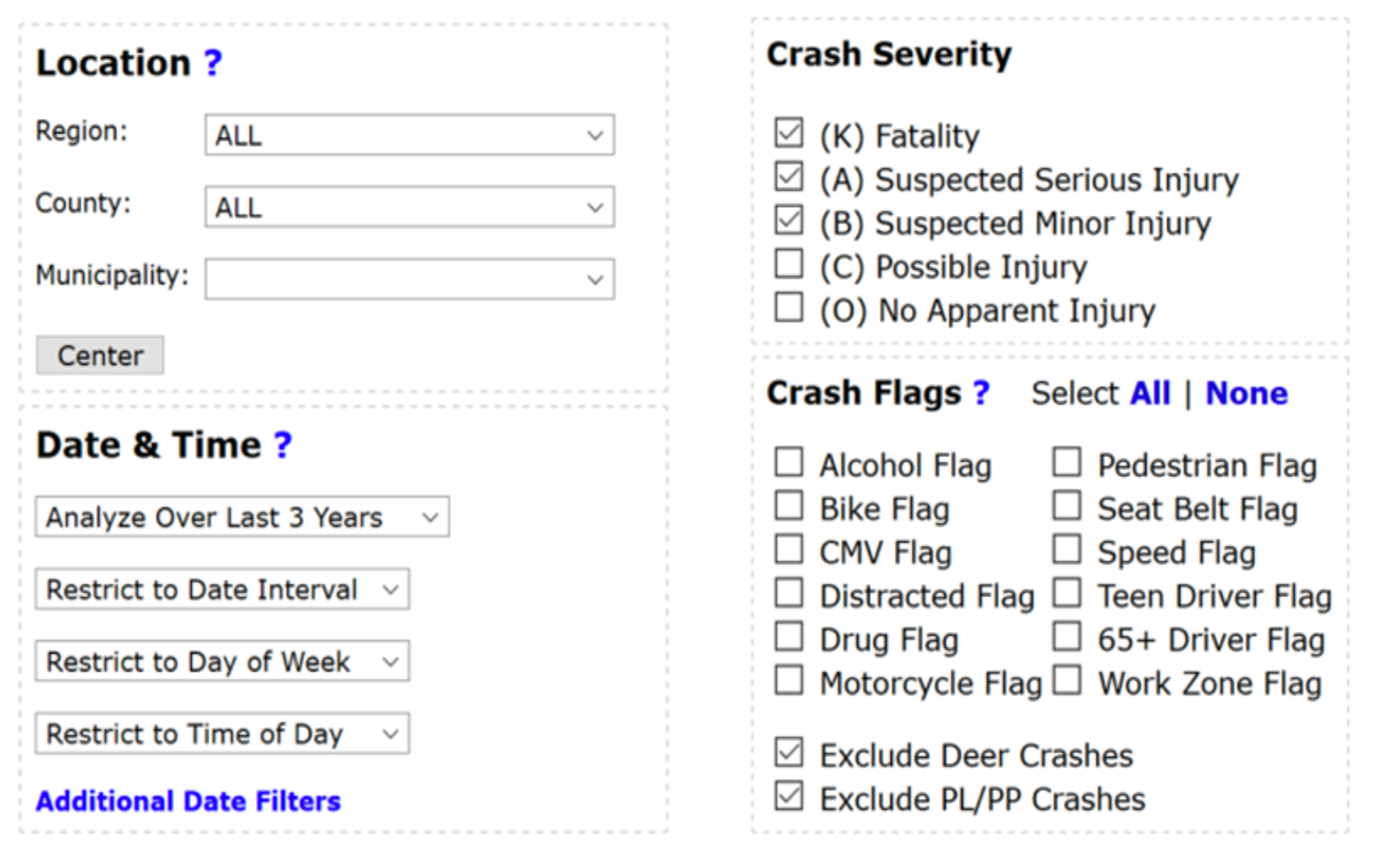}\\
    {\small \textbf{b} Filter Controls.}

    \vspace{0.5em}

    \includegraphics[width=\linewidth]{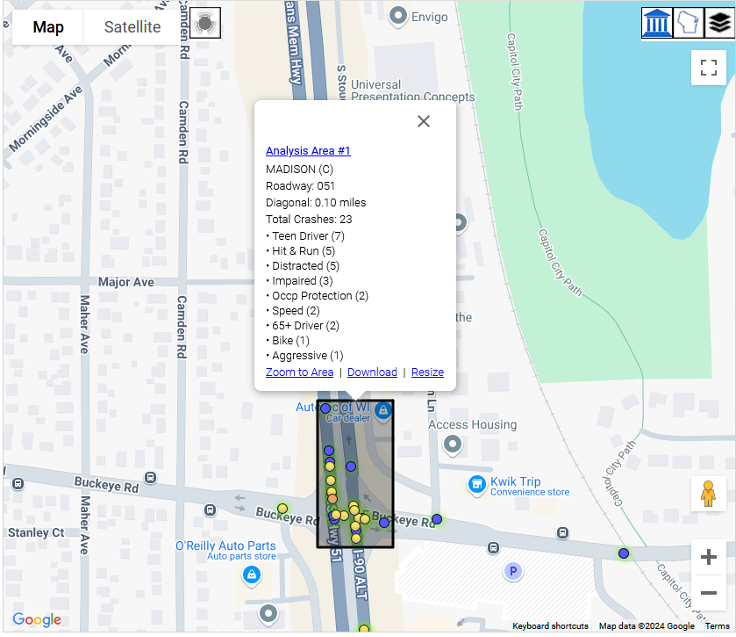}\\
    {\small \textbf{c} Hotspot Detail Popup.}
  \end{minipage}
  \caption{Community Maps Predictive Analytics interface: (a) spatial hotspot heatmap; (b) filter controls; (c) hotspot detail popup.}
  \label{fig:predictive_analytics_overview}
\end{figure}

Learned spatial--temporal models have sharpened short-term forecasting of urban events. Convolutional recurrent networks~\cite{shi2015convolutional}, spatio-temporal graph and attention architectures~\cite{yu2017spatio, vaswani2017attention, wu2026learning, gong2025integrating}, and self-exciting point processes~\cite{mohler2011self} capture local dynamics, network structure, and near-repeat clustering, and this machinery has been adapted directly to crashes, from real-time and Bayesian crash-risk models~\cite{hossain2012bayesian, theofilatos2019comparing, li2020real, li2022hybrid, wu2025ai2} to deep citywide accident-risk prediction~\cite{ren2018deep, yuan2018hetero, bao2019spatiotemporal, cheng2022work} and minute-level forecasting~\cite{zhou2020riskoracle}. These models raise accuracy, yet they optimize a single smoothed risk field and rarely deliver the severity-aware robustness, the explicit hotspot geometry, and the interpretable life-cycle structure that operational decisions require~\cite{tamakloe2025pattern, yeddula2023traffic}. More fundamentally, the underlying demands conflict. Minimizing grid-level error rewards smoothing, which erases the geometry that defines a hotspot; robustness in sparse rural counties competes with sensitivity in dense ones; and a single accurate forecast reveals nothing about temporal evolution, whereas practitioners need to know when a cluster emerges, grows, persists, declines, and vanishes~\cite{al2021measuring}. Such life-cycle accounts separate fleeting spikes from sustained threats and set the timing of intervention~\cite{mesic2024identifying, gui2024revealing}. No existing tool resolves this tension, delivering early detection, calibrated short-term forecasting, and an interpretable life cycle within one model that serves both dense and sparse regimes.

These capabilities have typically lived in separate systems: a clustering tool for detection, a forecaster for intensity, and a heuristic tracker for evolution. Splitting them fragments the signal and forces practitioners to reconcile inconsistent outputs. Our premise is that the three tasks share one substrate. A single spatial--temporal representation that encodes severity, cluster structure, cyclicity, and long-run exposure can drive detection, forecasting, and tracking together, given two design choices. First, the forecaster anchors its output on the county exposure map, which keeps predictions physically plausible and portable rather than freely regressed. Second, a lightweight expert mixture specializes the shared features to the local crash regime, so one statewide model serves both dense-urban and sparse-rural settings without per-location retraining. Anchoring and specialization together let a single model be accurate where data are rich and stable where they are scarce, while preserving the hotspot geometry that downstream tracking needs.

We realize this premise as \ours{} (Hotspot Emergence, Risk Anticipation, and Life-cycle Dynamics), a unified deep learning framework for crash-hotspot intelligence that delivers all three capabilities from a single statewide model:
\begin{enumerate}
  \item \textbf{Emerging-hotspot detection.} Multi-channel spatial--temporal grids encoding severity-stratified injuries, hotspot cluster structure, and temporal cyclicity, combined with a static exposure prior, feed a clustering pipeline and a dedicated emergence head that surface clusters of rising crash intensity before they consolidate into chronic hotspots.
  \item \textbf{Short-term forecasting.} A compact squeeze-and-excitation encoder with a factorized axial transformer emits a structured forecast that composes an exposure-anchored background, Hawkes-style self-excitation~\cite{mohler2011self}, and a zero-inflated negative-binomial intensity suited to over-dispersed crash counts~\cite{lord2010statistical}, together with explicit hotspot centroids. A regime-routed mixture-of-experts~\cite{shazeer2017outrageously} lets one statewide model specialize between dense-urban and sparse-rural regimes at under three percent additional parameters, and an error-aligned calibration recipe closes the gap between training objective and reported metrics.
  \item \textbf{Life-cycle tracking.} A minimal-cost matching scheme propagates cluster identities over time and labels each hotspot as \emph{birth}, \emph{growth}, \emph{stable}, \emph{decline}, or \emph{death}, yielding a compact, interpretable trajectory of risk evolution.
\end{enumerate}

We evaluate the full pipeline on a three-year crash corpus (2018--2020) from six Wisconsin counties spanning a wide range of urbanization, crash volume, and signal sparsity. Trained once statewide under a protocol shared with five learned baselines, \ours{} attains the best grid-level error while keeping hotspot localization competitive with the strongest baselines, and leads both on the densest county. It also flags emerging high-risk cells before they consolidate and produces temporally coherent life-cycle summaries, with the clearest gains in dense and moderate regimes. The remainder of the paper describes the study area and data (Section~\ref{sec:problem_statement}), formulates each module of \ours{} (Section~\ref{sec:methodology}), reports the evaluation (Section~\ref{sec:experiment}), and concludes with operational implications and future work (Section~\ref{sec:conclusion}).

\section{Problem Statement}\label{sec:problem_statement}

Our primary objective is to furnish traffic safety practitioners with an integrated toolkit that can (i) \emph{detect} emergent crash hotspots in near real time, (ii) \emph{forecast} the short-term spatial distribution and geometry of future crash clusters, and (iii) \emph{track} each hotspot's evolution through an interpretable life-cycle framework. In the subsections that follow, we first describe the data source and study area, then detail the temporal and spatial partitioning, and conclude with a high-level framing of the unified pipeline used to realize these goals.

\subsection{Data Selection and Study Area}

The crash data used in this study are sourced from the Community Maps Predictive Analytics system developed by the Traffic Operations and Safety (TOPS) Laboratory at the University of Wisconsin--Madison. The platform offers an interactive environment for querying, filtering, and inspecting crash records along several axes: spatial location, time, severity, and contributing factors. It supports both granular point-level exploration and aggregated hotspot summarization. Figure~\ref{fig:predictive_analytics_overview} illustrates the interface: panel (a) shows the spatial hotspot heatmap overlay used for rapid risk localization, (b) presents the filtering controls that tailor the dataset by region, temporal window, severity, and crash flags, and (c) displays a detailed hotspot popup with aggregated metrics for a selected area. These affordances are effective for retrospective inspection, but they remain static and query-driven; they neither anticipate where risk will concentrate next nor describe how a hotspot evolves. This gap motivates the more structured, predictive, and temporally coherent framework developed here.

\begin{figure}[pos=!t]
  \centering
  \begin{minipage}[b]{0.48\textwidth}
    \centering
    \includegraphics[width=\linewidth]{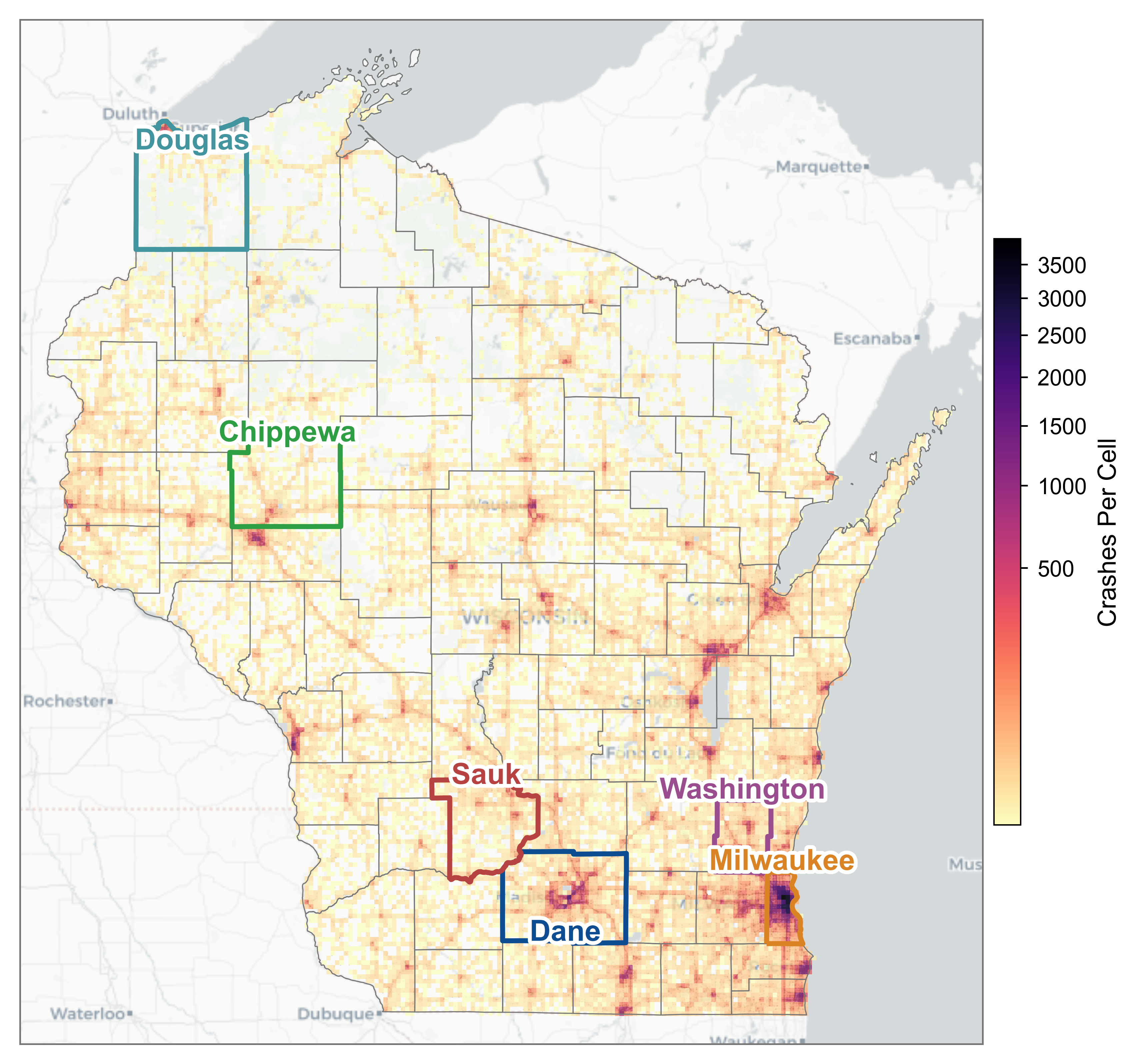}\\
    {\small \textbf{a} Statewide Crash Density.}
  \end{minipage}
  \hspace{0.4em}
  \begin{minipage}[b]{0.46\textwidth}
    \centering
    \includegraphics[width=\linewidth]{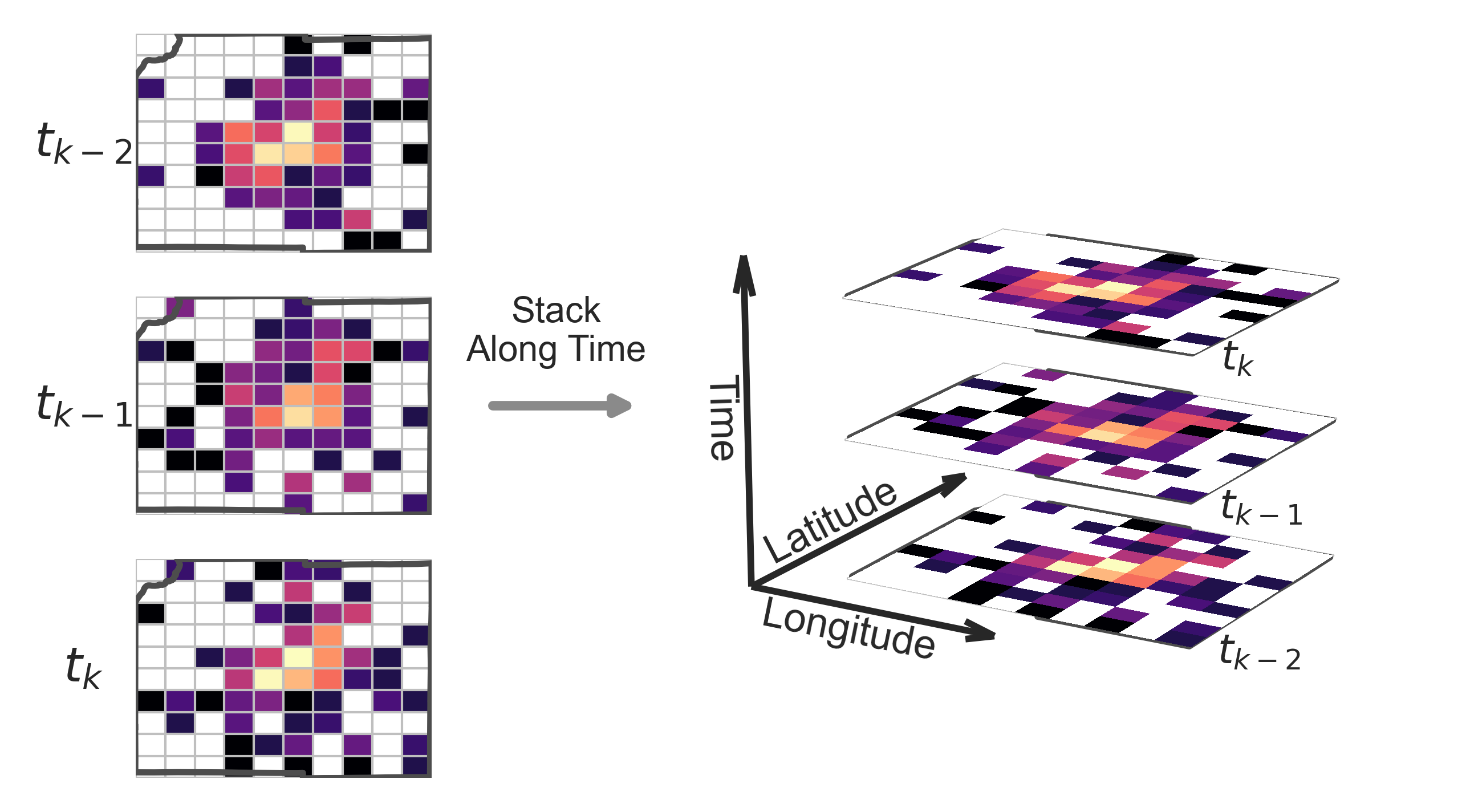}\\
    {\small \textbf{b} Building the Space--Time Cube (Dane County).}

    \vspace{0.6em}

    \includegraphics[width=0.80\linewidth]{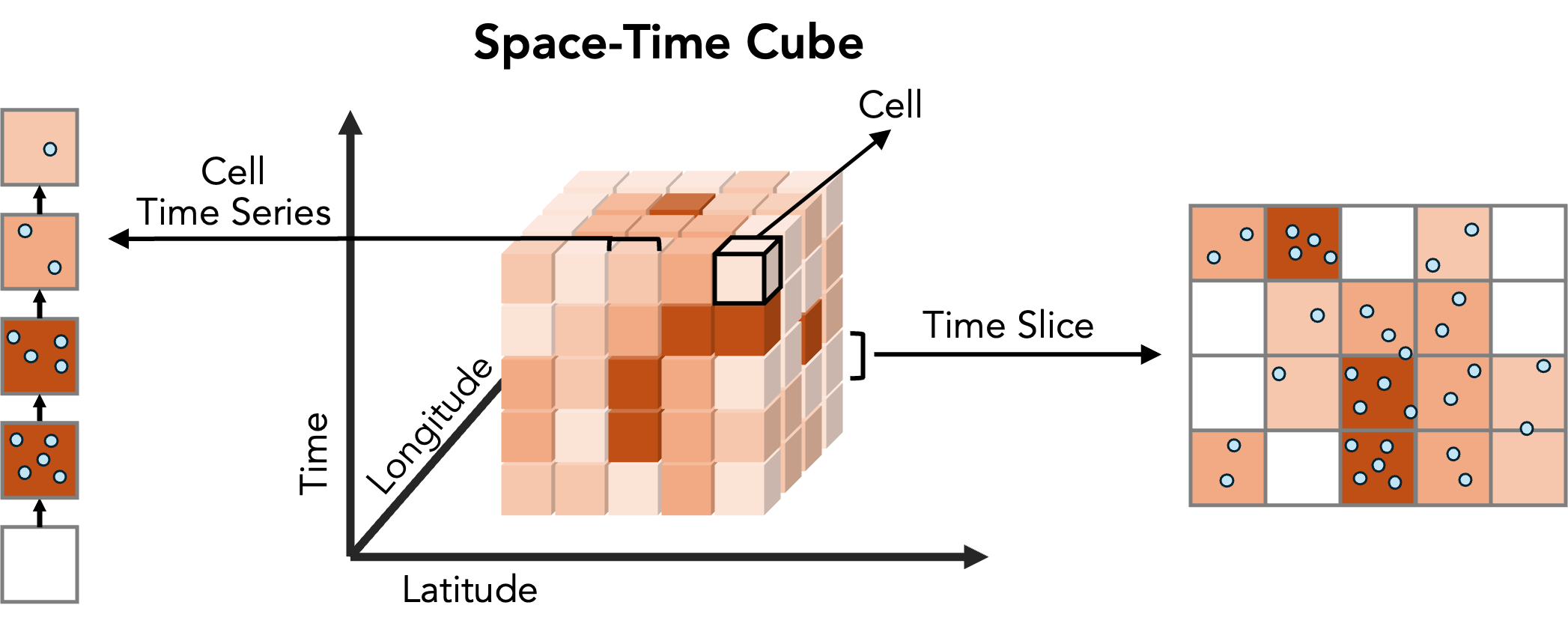}\\
    {\small \textbf{c} Space--Time Cube Schematic.}
  \end{minipage}
  \caption{Study area and spatial--temporal discretization: (a) statewide crash density (2018--2020) with the six study counties outlined in color; (b) weekly crash-injury grids stacked into a space--time cube for Dane County; (c) space--time cube schematic.}
  \label{fig:study_area_cube}
\end{figure}

We analyze all reported vehicle crashes in six Wisconsin counties, namely Dane, Sauk, Douglas, Washington, Chippewa, and Milwaukee, over the three-year period from January~1, 2018 through December~31, 2020. Each event carries geographic coordinates \((x_i,y_i)\), a precise timestamp \(t_i\), and a severity tier \(s_i \in \{\mathrm{K},\mathrm{A},\mathrm{B},\mathrm{C},\mathrm{O}\}\) for fatal, incapacitating, non-incapacitating, possible-injury, and property-damage-only outcomes. It also provides injury counts by severity, which enable the construction of impact scores, and binary crash-type flags such as \texttt{IMPAIRED}, \texttt{SPEEDFLAG}, \texttt{BIKEFLAG}, and \texttt{PEDESTRIANFLAG} that capture contributory context. Together, the spatial, temporal, and severity attributes underpin the detection and forecasting components of the framework. Figure~\ref{fig:study_area_cube}(a) situates the study geographically, showing statewide crash density over a basemap. Individual crash locations aggregate into a density surface that traces the road network, all county boundaries are drawn for reference, and the six focal counties are outlined in distinct colors. Their spatial dispersion and the diversity of crash regimes are evident, ranging from high-volume urbanized cores to lower-density rural corridors. These heterogeneities in geography and density directly inform our evaluation design and motivate the disaggregated, per-county analysis used throughout.

The six counties are chosen so that, together, they stress-test the pipeline across the full range of operating conditions the state presents; all reported crashes in each county over the study period enter the corpus, so the counties differ by their intrinsic crash regimes rather than by sampling design. Milwaukee County, the state's most populous, contributes by far the densest and highest-volume urban regime and the busiest road network in the study, and Dane County, the second-most populous, a second high-volume urban core. Washington County adds a suburban regime of intermediate density. Sauk and Douglas counties exemplify mixed urban--rural environments in which naive density-based hotspot methods tend to overemphasize urbanized pockets, and Chippewa County adds a further low-density rural regime with the sparsest signal. Collectively, the six counties span more than two orders of magnitude in crash density and more than an order of magnitude in weekly crash volume (Table~\ref{tab:weekly_crash_stats}), exposing the pipeline to the range of conditions it must serve in deployment.

\subsection{Temporal and Spatial Partitioning}

The study period is segmented into consecutive, non-overlapping windows of length \(\Delta T = 1\) week, yielding roughly \(T=156\) sequential windows per county indexed by their end-times \(\{t_1,\dots,t_T\}\). Within window \(t_j\), we retain all crashes satisfying
\begin{equation}
t_j - \Delta T < t_i \le t_j.
\end{equation}
This weekly discretization emphasizes recent trends, while the stacked history described below supplies sufficient context for robust clustering and short-term forecasting~\cite{Schubert2007}.

Figure~\ref{fig:study_area_cube}(b,c) depicts this temporal--spatial discretization as a space--time cube: successive spatial grid slices are stacked along the time axis, illustrating how historical context accumulates and is made available to downstream modules.

Spatially, each county's geographic extent is tessellated into a uniform grid of \(H\times W\) cells (with \(H=W=50\) in our implementation), formed by equal-width partitioning over the latitude and longitude ranges. Each cell \((u,v)\) thus corresponds to a consistent spatial footprint, enabling meaningful aggregation, comparison across regions, and efficient tensor operations. At forecast time \(t_{j+1}\), our dual objectives are: (1) to predict a \emph{grid-level intensity map} representing expected injury intensity per cell, and (2) to extract a set of explicit \emph{cluster centroids} \(\{\hat\mu_m\}\subset\mathbb R^2\) with associated severity summaries that capture the most salient emergent hotspots. Both predictions are conditioned on the preceding \(k=4\) windows \(\{t_{j-3}, t_{j-2}, t_{j-1}, t_j\}\), supplying temporal context for the detection and forecasting tasks.

\subsection{Unified Framework Overview}

We first state the learning problem. For each window \(t\), Section~\ref{sec:grid_construction} builds a multi-channel grid tensor \(\mathcal W_t\in\mathbb R^{C\times H\times W}\) that summarizes crash activity, and the \(k\) most recent windows are stacked into a history tensor \(\mathcal X_t=[\mathcal W_{t-k+1},\dots,\mathcal W_t]\in\mathbb R^{k\times C\times H\times W}\). Conditioned on \(\mathcal X_t\), a static per-county exposure map \(E_c\), and a county index \(c\), the framework predicts window \(t{+}1\) along two coupled objectives: (i) a \emph{grid-level intensity field} \(\widehat{\mathcal Y}_{t+1}\in\mathbb R^{5\times H\times W}\), the expected injury intensity in each cell and severity tier; and (ii) a set of \emph{hotspot descriptors}: a centroid heatmap \(\widehat Q_{t+1}\), a per-cell emergence probability \(\widehat B_{t+1}\), and a block-level change state \(\widehat O_{t+1}\), which together indicate where clusters will concentrate, where new ones will appear, and how existing ones are trending. We therefore learn a single mapping \(\mathcal F_\theta:(\mathcal X_t,E_c,c)\mapsto(\widehat{\mathcal Y}_{t+1},\,\widehat Q_{t+1},\,\widehat B_{t+1},\,\widehat O_{t+1})\), and evaluate its grid reconstruction and hotspot localization jointly. A downstream tracker then links clusters across windows and labels each with a life-cycle phase.

\begin{figure}[pos=!t]
  \centering
  \includegraphics[width=0.98\textwidth]{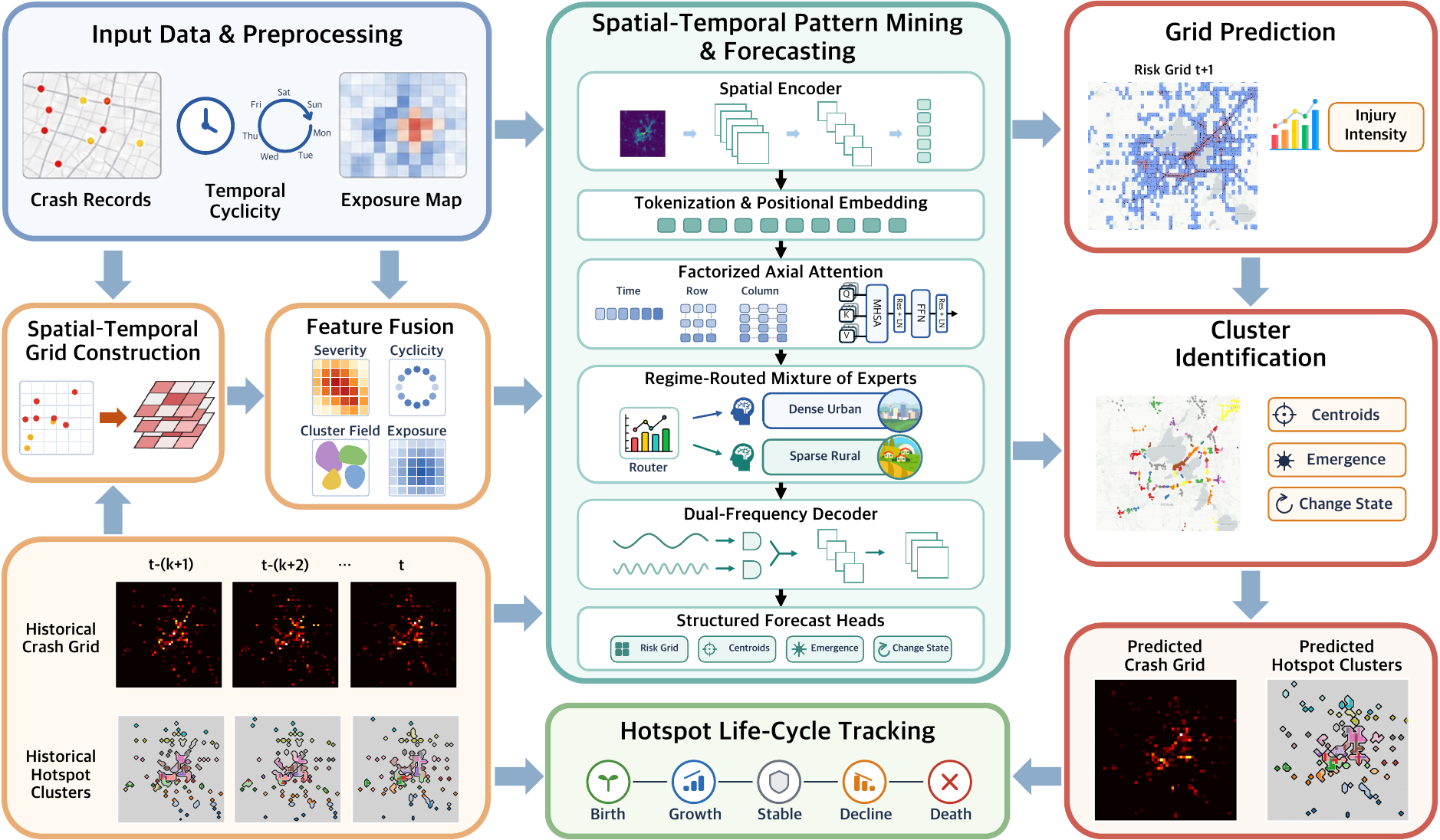}
  \caption{The unified \ours{} pipeline: (1) input preprocessing fuses crash records into multi-channel severity-stratified grid tensors; (2) the forecasting core predicts the next window's intensity map and hotspot centroids; (3) life-cycle tracking matches hotspots across windows and labels birth, growth, stable, decline, and death.}
  \label{fig:framework}
\end{figure}

Figure~\ref{fig:framework} consolidates this into a three-stage pipeline. The first stage, \emph{Input \& Preprocessing}, ingests raw crashes, aggregates severity-stratified injuries, and fuses them into the multi-channel grid tensors above. The second stage, \emph{Spatial--Temporal Pattern Mining \& Forecasting}, is the learned core of \ours{}: a squeeze-and-excitation convolutional encoder extracts local features from each history window, a factorized axial transformer propagates spatial--temporal context at reduced resolution, a regime-routed mixture-of-experts lets one statewide model specialize between dense urban and sparse rural regimes, and a structured output head composes an exposure-anchored background risk with a Hawkes-style self-excitation term under a zero-inflated negative-binomial likelihood, alongside dedicated centroid, emergence, and change-state heads. The third stage, \emph{Hotspot Life-Cycle Tracking}, combines historical and forecast grids with a severity-aware DBSCAN detector and a minimal-cost matching scheme to label life-cycle events (birth, growth, stable, decline, and death)~\cite{al2021measuring, gui2024revealing}. Together, these stages advance crash hotspot analysis from static retrospective mapping to proactive, interpretable, and predictive decision support. Section~\ref{sec:methodology} details the formulation of each module.

\section{Methodology}\label{sec:methodology}

We now detail the three core modules of \ours{} in turn: (i) spatial--temporal grid construction, which turns raw crash records into multi-channel tensors; (ii) the forecasting network, which predicts the next-window risk field and hotspot geometry; and (iii) hotspot life-cycle tracking, which converts those forecasts into interpretable trajectories. Each algorithmic operation is paired with its mathematical formulation and an intuitive account of its purpose. Notation is consistent throughout: \(t\) indexes discrete time windows, \(W\) and \(H\) are the grid resolutions in the longitudinal and latitudinal directions, yielding a \(W\times H\) tessellation, \(C\) is the number of input channels, and \(k\) the number of history slices.

\subsection{Spatial--Temporal Grid Construction}\label{sec:grid_construction}

To transform raw crash records into a format amenable to both clustering and deep forecasting, we discretize the study region and time into regularly sampled spatial--temporal tensors and compute multiple informative channels per spatial cell. For each time window \(t\), this process produces a multi-channel tensor \(\mathcal W_t\in\mathbb R^{C\times H\times W}\), where \(H\) and \(W\) index latitude and longitude bins, respectively, and \(C\) aggregates heterogeneous feature modalities.

\subsubsection{Spatial Discretization}
Let the longitudinal extent of the study area be \([x_{\min},x_{\max}]\) and the latitudinal extent \([y_{\min},y_{\max}]\). We partition these ranges into \(W\) and \(H\) equal-width intervals, yielding spatial resolutions
\begin{equation}
\Delta_x=\frac{x_{\max}-x_{\min}}{W},\quad
\Delta_y=\frac{y_{\max}-y_{\min}}{H}.
\end{equation}
Each grid cell is indexed by \((u,v)\) with \(u\in\{1,\dots,W\}\), \(v\in\{1,\dots,H\}\), and covers the geographic rectangle
\begin{equation}
x\in\bigl[x_{\min}+(u-1)\,\Delta_x,\;x_{\min}+u\,\Delta_x\bigr),\quad
y\in\bigl[y_{\min}+(v-1)\,\Delta_y,\;y_{\min}+v\,\Delta_y\bigr).
\end{equation}
This uniform tessellation preserves spatial locality and facilitates efficient tensor operations while providing a consistent reference frame across time.

\subsubsection{Intensity Aggregation}
Our input intensity aggregates reported injuries rather than weighted crash counts. Let \(\mathcal C_t\) be the set of crashes falling within window \(t\), and denote each crash \(i\in\mathcal C_t\) by its location \((x_i,y_i)\), occurrence time \(t_i\), its most severe injury tier \(s_i\in\{\mathrm{K,A,B,C,O}\}\), and the injury count \(n_i\) reported at that tier. For each severity tier \(s\), a cell-wise intensity channel accumulates the injuries of crashes of that tier falling into cell \((u,v)\):
\begin{equation}
W^s_t(u,v)
=
\sum_{\substack{i\in\mathcal C_t\\s_i=s\\(x_i,y_i)\in(u,v)}} n_i.
\end{equation}
This produces five \emph{severity intensity channels} \(\{W^K_t, W^A_t, W^B_t, W^C_t, W^O_t\}\) that together form a fine-grained representation of crash harm stratified by severity. Because property-damage-only crashes involve no injuries, the O channel carries no mass in practice and the effective signal is injury-based. Before modeling, each county's severity channels are normalized by a county scale \(s_c\), the mean positive cell intensity over that county's training windows, so that one model can serve counties of very different crash volumes.

\subsubsection{Temporal Cyclicity Channels}
Crash occurrences are influenced by recurring temporal patterns such as time of day and day of week. To encode such periodicities while preserving differentiability, we compute sinusoidal embeddings for hour-of-day and weekday at the cell level. For each cell \((u,v)\), let \(h_i\in\{0,\dots,23\}\) and \(d_i\in\{1,\dots,7\}\) be the hour and weekday of crash \(i\in\mathcal C_t\) whose location falls in that cell. Define \(N^t_{u,v}=\bigl|\{i\in\mathcal C_t:(x_i,y_i)\in(u,v)\}\bigr|\) as the number of window-\(t\) crashes in cell \((u,v)\). The average cyclic embeddings are then
\begin{equation}
E^h_t(u,v)
=
\frac{1}{N^t_{u,v}}
\sum_{i\in\mathcal C_t:(x_i,y_i)\in(u,v)}
\bigl(\cos\tfrac{2\pi h_i}{24},\;\sin\tfrac{2\pi h_i}{24}\bigr),
\quad
E^d_t(u,v)
=
\frac{1}{N^t_{u,v}}
\sum_{i\in\mathcal C_t:(x_i,y_i)\in(u,v)}
\bigl(\cos\tfrac{2\pi d_i}{7},\;\sin\tfrac{2\pi d_i}{7}\bigr),
\end{equation}
with cells containing no crashes assigned zero vectors. These four scalar channels, two for hour and two for weekday, capture periodic behavior such as rush hours, weekend effects, and daily traffic rhythms that modulate crash likelihood beyond raw counts.

\subsubsection{Cluster-Structure and Exposure Channels}
Two further channels inject hotspot structure and long-run spatial context. First, a \emph{Gaussian cluster field} summarizes the window's discrete hotspot geometry as a continuous surface. The window's summed intensity map is clustered with DBSCAN (Section~\ref{sec:life_cycle}), and the clustered intensity is diffused with an isotropic Gaussian kernel of bandwidth \(\sigma=2\) cells:
\begin{equation}
F_t(u,v)
=
\sum_{(u',v')\in\mathcal H_t}
G_t(u',v')\,
\exp\!\left(-\frac{(u-u')^2+(v-v')^2}{2\sigma^2}\right),
\end{equation}
where \(\mathcal H_t\) is the set of cells assigned to a cluster and \(G_t\) is the summed intensity. Compared with a raw integer cluster-label raster, this field representation carries cluster mass and extent in a form a convolutional encoder can exploit.

Second, a static \emph{exposure map} encodes each county's long-run spatial crash propensity at the scale of occupied-cell intensity. Let \(\bar W^s_c(u,v)\) denote the unconditional per-window mean of \(W^s_t(u,v)/s_c\) over the county's training windows \(\mathcal T_c\), and let \(\rho_c(u,v)\) be the ratio of the conditional-on-positive mean to the unconditional mean of the summed intensity in that cell. Because \(\rho_c\) is noisy where positive weeks are rare, it is shrunk toward its county median \(\tilde\rho_c\) and clipped:
\begin{equation}
\rho'_c = \mathrm{clip}\!\left(\tilde\rho_c\sqrt{\rho_c/\tilde\rho_c},\;
[0.5,\,2]\cdot\tilde\rho_c\right),
\qquad
E_c(u,v,s)
=
\max\!\bigl(\bar W^s_c(u,v)\,\rho'_c(u,v),\;\epsilon\bigr),
\qquad \epsilon = 0.05,
\label{eq:exposure}
\end{equation}
computed on training windows only so that no future information leaks into evaluation. Anchoring at the conditional scale matters: an unconditional-mean anchor sits far below occupied-cell intensity, forcing the bounded multiplicative correction of Section~\ref{sec:forecasting} into saturation and flattening the predicted risk surface. The channel-summed unconditional map \(\log\!\bigl(1+\sum_s \bar W^s_c\bigr)\), together with the per-cell occupancy rate, feeds a small static context branch in the decoder; the full map \(E_c\) anchors the forecast head as a multiplicative offset. Because crashes concentrate on the road network, this map plays the role of the exposure term in classical safety performance functions while requiring no auxiliary road inventory. We stress that it is a crash-history-based proxy rather than a true exposure measure such as traffic volume or vehicle-miles traveled: built solely from training-window crash intensities, it blends exposure with underlying risk.

\subsubsection{Channel Stacking and Final Tensor}
Collecting all channels yields the multi-channel grid tensor
\begin{equation}
\mathcal W_t
=
\bigl[\,W^K_t,\,W^A_t,\,W^B_t,\,W^C_t,\,W^O_t;\;F_t;\;E^h_t,\,E^d_t\,\bigr]
\;\in\;\mathbb R^{C\times H\times W},
\quad
C = 5 + 1 + 4 = 10.
\end{equation}
This tensor simultaneously encodes severity-stratified crash intensity, hotspot cluster structure, and temporal regularities; the long-run exposure context enters the network separately through the static branch of Section~\ref{sec:forecasting}.

\subsubsection{Temporal Stacking}
To leverage recent dynamics, we assemble the input for forecasting by concatenating the most recent \(k\) window tensors:
\begin{equation}
\mathcal X_t
=
\bigl[\mathcal W_{t-k+1},\,\mathcal W_{t-k+2},\,\dots,\,\mathcal W_t\bigr]
\;\in\;\mathbb R^{k\times C\times H\times W},
\quad k=4.
\end{equation}
Each slice \(\mathcal W_{t-\kappa}\) preserves spatial locality and captures severity-stratified intensities, cluster structure, and temporal cyclicity, yielding a rich, structured representation for downstream forecasting and life-cycle clustering.

\subsection{Spatial--Temporal Forecasting}\label{sec:forecasting}

\begin{figure}[pos=!t]
  \centering
  \includegraphics[width=\textwidth]{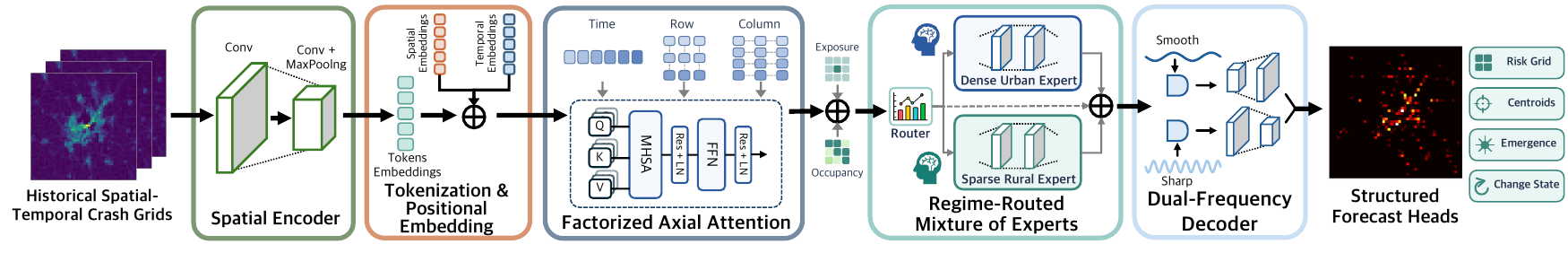}
  \caption{The \ours{} forecasting architecture: a squeeze-and-excitation encoder summarizes each history window, factorized axial attention mixes context along rows, columns, and time, a regime-routed mixture-of-experts specializes cells by density regime, and a dual-frequency decoder feeds the structured heads for exposure-anchored severity risk, centroids, emergence, and change states.}
  \label{fig:model_architecture}
\end{figure}

This module learns the parametric forecaster \(\mathcal F_{\theta}\) that maps the history tensor \(\mathcal X_t\in\mathbb R^{k\times C\times H\times W}\) to a structured prediction for window \(t{+}1\). The computation proceeds in five stages (Figure~\ref{fig:model_architecture}): (1) a weight-shared \emph{spatial encoder} that summarizes each history slice; (2) a \emph{factorized axial transformer} that mixes spatial and temporal context at reduced resolution; (3) a \emph{static-context branch} with per-county feature modulation; (4) a \emph{regime-routed mixture-of-experts} that specializes the shared features by local crash regime; and (5) a \emph{structured intensity head} with three auxiliary heads. Rather than regress intensity directly, the head decomposes the predicted severity field \(\boldsymbol\mu_{t+1}\) into a slowly varying \emph{background risk}, anchored multiplicatively on the county exposure map, and a \emph{self-excitation} term triggered by recent crashes, in the spirit of spatio-temporal point processes.

The forecaster emits five outputs for window \(t{+}1\): (i) the severity-stratified intensity \(\boldsymbol\mu_{t+1}\in\mathbb R^{5\times H\times W}\); (ii) a cell-occupancy logit \(p_{t+1}\in\mathbb R^{H\times W}\) that gates the zero-inflated intensity; (iii) a hotspot-centroid heatmap \(\widehat Q_{t+1}\in[0,1]^{H\times W}\); (iv) a per-cell emergence (birth) probability \(\widehat B_{t+1}\in[0,1]^{H\times W}\) for historically quiet cells; and (v) an ordinal, block-level change state \(\widehat O_{t+1}\in\mathbb R^{4\times\frac{H}{5}\times\frac{W}{5}}\) over \{cooling, stable, warming, emerging\}. The per-cell injury field reported at evaluation is the zero-inflated expectation \(\mathbb E[\widehat{\mathcal Y}^{s}_{t+1}]=\sigma(p_{t+1})\,\boldsymbol\mu^{s}_{t+1}\,s_c\), where \(\sigma\) is the sigmoid and \(s_c\) the county scale. We describe the five stages and the output heads in turn, and state the training objective in Section~\ref{sec:losses}.

\subsubsection{Spatial Encoder}

The model begins by encoding each of the \(k\) historical grid slices \(\mathcal W_{t-k+\kappa}\) into localized latent representations. For each \(\kappa\), we apply two convolutional blocks with intermediate pooling, each followed by a squeeze-and-excitation (SE) channel recalibration:
\begin{equation}
\begin{aligned}
Z_\kappa^{(1)}
&= \mathrm{SE}\!\left(\mathrm{ReLU}\!\left(\mathrm{BN}\!\left(\mathrm{Conv2D}_{C\to16}(\mathcal W_{t-k+\kappa})\right)\right)\right),\\
Z_\kappa^{(2)}
&= \mathrm{SE}\!\left(\mathrm{ReLU}\!\left(\mathrm{BN}\!\left(\mathrm{Conv2D}_{16\to32}\!\left(\mathrm{MaxPool2D}(Z_\kappa^{(1)},2)\right)\right)\right)\right),
\end{aligned}
\end{equation}
producing feature maps \(Z_{\kappa}^{(2)}\in\mathbb R^{32\times \frac{H}{2}\times \frac{W}{2}}\). Here \(\mathrm{Conv2D}_{a\to b}\) denotes a \(3\times3\) convolution with \(a\) input and \(b\) output channels (stride 1, padding 1), \(\mathrm{MaxPool2D}(\cdot,2)\) halves spatial dimensions, and \(\mathrm{SE}\) rescales channels by a learned global gate. The deliberately slim channel widths of 16 and 32 reflect the small-sample regime: wider encoders were found to overfit before exploiting their capacity. The full-resolution features \(Z^{(1)}_{k}\) of the most recent window are retained as a skip path for the sharpening branch described below.

\subsubsection{Tokenization and Positional Encoding}

To enable explicit spatial--temporal attention, each downsampled feature map \(Z_{\kappa}^{(2)}\) is flattened along its spatial dimensions into \(L = \frac{H}{2}\times\frac{W}{2}\) tokens and projected into the attention space:
\begin{equation}
T_{\kappa}
= \mathrm{FlattenSpatial}(Z_{\kappa}^{(2)})\,W_{\mathrm{proj}}
+ P_{\mathrm{spatial}}
+ P_{\mathrm{temp}}^{(\kappa)},
\end{equation}
where \(W_{\mathrm{proj}}\in\mathbb R^{32\times d}\) lifts tokens to the latent dimension \(d=96\). \(P_{\mathrm{spatial}}\in\mathbb R^{L\times d}\) and \(P_{\mathrm{temp}}^{(\kappa)}\in\mathbb R^{d}\) are \emph{fixed sinusoidal} encodings of the token's cell index and history step. Fixed encodings are preferred over learned positional tables, which proved to be a consistent source of overfitting at this sample size.

\subsubsection{Factorized Axial Spatial--Temporal Attention}

Dense attention over all \(kL\) tokens is both quadratic in cost and prone to overfitting in a data-sparse regime. We therefore factorize each attention layer along the grid axes: multi-head self-attention (MHA) is applied first along each row of the downsampled grid, then along each column, and finally across the \(k\) history steps for each cell. Writing the token tensor as \(T\in\mathbb R^{k\times \frac{H}{2}\times \frac{W}{2}\times d}\), one axial layer computes
\begin{equation}
\begin{aligned}
T &\leftarrow \mathrm{LN}\bigl(T + \mathrm{MHA}_{\mathrm{row}}(T)\bigr),\\
T &\leftarrow \mathrm{LN}\bigl(T + \mathrm{MHA}_{\mathrm{col}}(T)\bigr),\\
T &\leftarrow \mathrm{LN}\bigl(T + \mathrm{MHA}_{\mathrm{time}}(T)\bigr),\\
T &\leftarrow \mathrm{LN}\bigl(T + \mathrm{FFN}(T)\bigr),
\end{aligned}
\end{equation}
Here \(\mathrm{LN}\) is layer normalization and \(\mathrm{FFN}\) a two-layer position-wise network of hidden width \(2d\). Each \(\mathrm{MHA}\) uses \(n_h=4\) heads of the standard scaled dot-product form: from projections \((Q,K,V)=(TW_Q,TW_K,TW_V)\),
\begin{equation}
\mathrm{MHA}(T)=\bigl[\,\mathrm{head}_1,\dots,\mathrm{head}_{n_h}\,\bigr]\,W_O,
\qquad
\mathrm{head}_i=\mathrm{softmax}\!\Bigl(\tfrac{Q_i K_i^{\!\top}}{\sqrt{d/n_h}}\Bigr)V_i,
\label{eq:mha}
\end{equation}
applied \emph{independently along one axis at a time}: \(\mathrm{MHA}_{\mathrm{row}}\) attends over the \(W/2\) tokens within each grid row, \(\mathrm{MHA}_{\mathrm{col}}\) over the \(H/2\) tokens within each column, and \(\mathrm{MHA}_{\mathrm{time}}\) over the \(k\) history steps at each cell. Chaining the three axes gives every cell a full spatio-temporal receptive field within a single layer while reducing the attention cost from \(\mathcal O\!\bigl((kL)^2\bigr)\) for dense attention to \(\mathcal O\!\bigl(kL(\tfrac{H}{2}+\tfrac{W}{2}+k)\bigr)\), a decisive saving at this sample size. We stack \(L_a=2\) such layers. Collecting the per-step tokens \(T_\kappa\) into a token tensor \(T^{(0)}=\operatorname*{Stack}_{\kappa=1}^{k}\!\bigl(T_\kappa\bigr)\in\mathbb R^{k\times\frac H2\times\frac W2\times d}\) and iterating \(T^{(\ell)}=\mathrm{AxialLayer}\bigl(T^{(\ell-1)}\bigr)\), the most-recent history step of the final tensor is read out and reshaped into the coarse feature map that feeds the decoder,
\begin{equation}
z_t = \operatorname{Reshape}\!\bigl(T^{(L_a)}_{k}\bigr)\in\mathbb R^{d\times\frac{H}{2}\times\frac{W}{2}}.
\label{eq:zt}
\end{equation}

\subsubsection{Static Context and Coarse Feature Fusion}

A small \emph{static-context branch} encodes the two time-invariant maps \(S=[\,s^{\mathrm{exp}},\,s^{\mathrm{occ}}\,]\in\mathbb R^{2\times H\times W}\), where \(s^{\mathrm{exp}}=\log(1+\sum_s\bar W^s_c)\) is the log unconditional exposure heat and \(s^{\mathrm{occ}}\) is the per-cell occupancy rate (Section~\ref{sec:grid_construction}), with two \(3\times3\) convolutions, producing an eight-channel static feature map \(\bar S\). Placing exposure and occupancy in this separate branch keeps the temporal encoder driven by crash dynamics alone, while long-run spatial structure still supports decoding. Pooled to the working resolution and concatenated with the transformer's most-recent-step output \(z_t\in\mathbb R^{d\times\frac H2\times\frac W2}\), it forms the fused coarse representation
\begin{equation}
u = \bigl[\,z_t;\ \bar S\,\bigr]\ \in\ \mathbb R^{(d+8)\times\frac H2\times\frac W2}.
\end{equation}
A \(3\times3\) convolution maps \(u\) to a \(32\)-channel coarse feature. Before it is county-modulated and decoded, this feature is refined per crash regime by the regime-routed mixture-of-experts of the next subsection.

\subsubsection{Regime-Routed Mixture-of-Experts Refinement}

Crash generation in a dense urban core and in a sparse rural corridor follows visibly different regimes: overlapping, weekly-shifting hotspots in the former, and rare, threshold-critical events in the latter. Yet a single statewide network must serve both. To let shared weights specialize without duplicating the model, the coarse feature map is refined by a lightweight regime-routed mixture-of-experts: two experts routed by the local density regime. Writing \(u\) for the concatenation of the attention output and the pooled static features, the refined representation is
\begin{equation}
f = \mathrm{Conv}(u)
+ g_{\mathrm{dense}}(u)\odot E_{\mathrm{dense}}(u)
+ g_{\mathrm{sparse}}(u)\odot E_{\mathrm{sparse}}(u),
\label{eq:moe}
\end{equation}
where \(\mathrm{Conv}(u)\) is the shared coarse convolution of the previous subsection and each expert is a depthwise-separable convolution with a \emph{zero-initialized} output layer,
\begin{equation}
E_i(u) = \mathrm{Conv}_{1\times1}\!\bigl(\mathrm{ReLU}(\mathrm{DWConv}_{3\times3}(u))\bigr),
\qquad i\in\{\mathrm{dense},\,\mathrm{sparse}\}.
\label{eq:expert}
\end{equation}
The zero initialization makes the mixture reproduce the plain backbone exactly at the start of training; the experts depart from it only where the data justify specialization. The two cell-wise gates are produced by a two-layer \(1\times1\) convolutional router \(R\) that reads the density regime directly off the static exposure and occupancy maps (pooled to the working resolution, \(s^{\mathrm{exp}}_{\downarrow},s^{\mathrm{occ}}_{\downarrow}\)) and the county embedding \(e_c\):
\begin{equation}
\bigl(g_{\mathrm{dense}}(u),\,g_{\mathrm{sparse}}(u)\bigr)
= \mathrm{softmax}\!\Bigl(R\bigl(\bigl[\,s^{\mathrm{exp}}_{\downarrow};\ s^{\mathrm{occ}}_{\downarrow};\ e_c\,\bigr]\bigr)\Bigr),
\qquad g_{\mathrm{dense}}+g_{\mathrm{sparse}}=1 .
\label{eq:moe_gate}
\end{equation}
Because these inputs \emph{define} the regime, the router needs no auxiliary supervision or load-balancing loss, and in training we observed no collapse: dense urban cells route consistently toward one expert and sparse rural cells toward the other, with soft blends in transitional areas. The whole mixture of two experts and the router adds roughly \(9\)k parameters, under 3\% of the model, preserving the small-sample capacity discipline of the backbone.

\subsubsection{County Modulation and Dual-Frequency Decoder}

The refined coarse feature \(f\) of Equation~\eqref{eq:moe} is modulated per county with feature-wise linear modulation (FiLM): the county embedding \(e_c\) produces channel-wise scale and shift vectors that adapt one statewide model to county-specific dynamics,
\begin{equation}
[\,\boldsymbol\gamma_c;\ \boldsymbol\beta_c\,] = W_{\mathrm{film}}\,e_c + b_{\mathrm{film}},
\qquad
\tilde f = f\odot(1+\boldsymbol\gamma_c) + \boldsymbol\beta_c,
\label{eq:film}
\end{equation}
with \((\boldsymbol\gamma_c,\boldsymbol\beta_c)\) broadcast over the coarse grid. Two branches then decode \(\tilde f\) at complementary frequencies. A \emph{smooth} branch applies a large-kernel (\(7\times7\)) convolution at the coarse resolution and upsamples bilinearly to capture low-frequency risk structure; a \emph{sharp} branch fuses the upsampled features with the full-resolution encoder skip path \(Z^{(1)}_{k}\) and the static features \(\bar S\), then applies a \(3\times3\) head to restore cell-level detail,
\begin{equation}
\mathrm{smooth} = \mathrm{Up}_{H\times W}\!\bigl(\mathrm{Conv}_{7\times7}(\tilde f)\bigr),
\qquad
\mathrm{sharp} = \mathrm{Conv}_{3\times3}\!\Bigl(\mathrm{Fuse}\!\bigl[\,\mathrm{Up}_{H\times W}(\tilde f);\ Z^{(1)}_{k};\ \bar S\,\bigr]\Bigr).
\label{eq:smoothsharp}
\end{equation}
Both branches output five severity channels and are summed to form the log-risk modulation:
\begin{equation}
g_{t+1}
=
\mathrm{smooth}
+
\mathrm{sharp},
\label{eq:gated}
\end{equation}
where \(g_{t+1}\in\mathbb R^{5\times H\times W}\) is the log-risk modulation driving the intensity head below, where a saturating \(\tanh\) bounds its effect. The occupancy logit \(p_{t+1}\) is not used to gate this sum; it enters the model only through the zero-inflated expectation and the occupancy loss, so the sharp detail is preserved everywhere rather than suppressed off the occupancy mask.

\subsubsection{Structured Intensity Head: Exposure Background and Self-Excitation}

The predicted severity intensity combines two nonnegative components. The \emph{background} term modulates the county exposure map of Equation~\eqref{eq:exposure} multiplicatively in log space,
\begin{equation}
\boldsymbol\mu^{\mathrm{bg}}_{t+1}
=
\exp\!\bigl(\log E_c + b + b_c + 4\tanh(g_{t+1}/4)\bigr)
=
E_c\,\exp\!\bigl(b + b_c + 4\tanh(g_{t+1}/4)\bigr),
\label{eq:background}
\end{equation}
so the network learns \emph{deviation from historical risk} rather than absolute intensity; the bounded exponent keeps the modulation within a factor of \(e^{\pm4}\). The exposure map enters with unit coupling, meaning the anchor exponent is fixed at \(1\). The background is then an exact multiplicative offset on the exposure \(E_c\), that is, full exposure anchoring rather than a learned power. The global scalar \(b\) absorbs any residual gain, and a per-county scalar \(b_c\) lets each county calibrate its own amplitude, which proved essential for the sparsest counties. Fixing the exponent rather than learning it simplifies the head at no cost to accuracy. This mirrors the multiplicative exposure structure of classical safety performance functions and makes the learned component portable across the study counties. The \emph{self-excitation} term captures the empirical regularity that recent crashes elevate near-future risk nearby. With \(h_{t-\kappa}\) denoting the summed severity map of history step \(\kappa\),
\begin{equation}
\boldsymbol\mu^{\mathrm{ex}}_{t+1}
=
\Bigl(\sum_{\kappa=1}^{k} w_\kappa\,\bigl(K \ast h_{t+1-\kappa}\bigr)\Bigr)\otimes \boldsymbol\alpha,
\qquad
w_\kappa\ge0,
\label{eq:hawkes}
\end{equation}
where \(K\) is a single learned nonnegative \(9\times9\) spatial kernel shared across lags, \(w_\kappa\) are per-lag weights scaled by a learned global gain, and \(\boldsymbol\alpha\in\Delta^{4}\) is a learned softmax allocation over the five severity channels. This branch is a discretized Hawkes triggering kernel with roughly one hundred parameters. The final intensity and its evaluation-time expectation are
\begin{equation}
\boldsymbol\mu_{t+1} = \boldsymbol\mu^{\mathrm{bg}}_{t+1} + \boldsymbol\mu^{\mathrm{ex}}_{t+1},
\qquad
\mathbb E[\widehat{\mathcal Y}^{s}_{t+1}] = \sigma(p_{t+1})\odot\boldsymbol\mu^{s}_{t+1}\cdot s_c .
\end{equation}

\subsubsection{Auxiliary Output Heads}

Three further \emph{output heads} share the sharp-branch features and, together with the intensity field \(\boldsymbol\mu_{t+1}\) and the occupancy logit \(p_{t+1}\), complete the model's outputs. The \emph{centroid head} outputs a hotspot-center heatmap \(\widehat Q_{t+1}\in[0,1]^{H\times W}\), supervised by Gaussian-rendered ground-truth centroids from the cluster extraction of Section~\ref{sec:life_cycle}. The \emph{emergence (birth) head} outputs a per-cell probability \(\widehat B_{t+1}\in[0,1]^{H\times W}\) that a historically quiet cell (no crashes in the \(k\) input windows) becomes active at or above one injury, an explicit early-warning product. The \emph{change-state head} average-pools the features to \(5\times5\)-cell blocks and outputs class scores \(\widehat O_{t+1}\in\mathbb R^{4\times\frac{H}{5}\times\frac{W}{5}}\) over \{cooling, stable, warming, emerging\} relative to the recent input windows, a standalone block-level summary of local trend. These three, with \(\boldsymbol\mu_{t+1}\) and \(p_{t+1}\), are exactly the five outputs of \(\mathcal F_\theta\) declared at the start of Section~\ref{sec:forecasting}. The centroid, emergence, and change-state heads act as auxiliary supervision and standalone operational products; the life-cycle tracker of Section~\ref{sec:life_cycle} operates on the intensity forecasts alone.

\subsubsection{Loss Functions}\label{sec:losses}

The training objective splits into two groups: a \emph{structured intensity loss} \(\mathcal L_{\mathrm{int}}\) that supervises the forecast \(\boldsymbol\mu_{t+1}\) and its occupancy gate, and an \emph{auxiliary loss} \(\mathcal L_{\mathrm{aux}}\) that supervises the three side heads. Throughout, \(\mathcal Y_{t+1}\in\mathbb R^{5\times H\times W}\) denotes the observed severity-intensity target, the five severity channels of window \(t{+}1\) in county-normalized units, with \(\mathcal Y_{t+1}(u,v)\) its severity-summed value at cell \((u,v)\); \(y_{t+1}=s_c\,\mathcal Y_{t+1}\) is the same target in raw injury units. We define each term below, then combine the two groups.

The intensity group comprises three complementary terms: an occupancy gate, a count likelihood, and a calibration tie. Cell occupancy is trained with binary cross-entropy against the indicator \(O_{t+1}(u,v)=\mathbb 1[\mathcal Y_{t+1}(u,v)>0]\) of nonzero observed intensity, forming the zero-inflation gate:
\begin{equation}
\mathcal L_{\mathrm{occ}}
= -\frac{1}{HW}\sum_{u,v}\Bigl[\,O_{t+1}\log\sigma(p_{t+1}) + (1-O_{t+1})\log\bigl(1-\sigma(p_{t+1})\bigr)\Bigr].
\end{equation}
On occupied cells, intensities are trained in raw injury units with a negative binomial (NB) likelihood, the classical choice for over-dispersed crash counts~\cite{abdel2000modeling, lord2005poisson, park2007multivariate}, applied independently to each severity channel with a single shared dispersion,
\begin{equation}
\mathcal L_{\mathrm{NB}}
=
-\frac{1}{|\Omega|}\sum_{(u,v)\in\Omega}
\log \mathrm{NB}\!\bigl(y_{t+1}\,;\,\mu_{t+1}\,s_c,\;r\bigr),
\qquad
\Omega=\{(u,v):O_{t+1}=1\},
\end{equation}
whose mean is \(\mu_{t+1}s_c\) and whose learned dispersion \(r\), converging to \(r\approx3.6\), captures the substantial overdispersion that a squared-error or Tweedie objective mishandles. Strictly, this occupancy--count factorization is a hurdle-style two-part scheme rather than a mixture-form zero-inflated NB likelihood: all zeros are carried by the gate and the NB term is fit on occupied cells only. We retain the zero-inflated shorthand for its familiarity in the crash-frequency literature. Because occupancy and count are trained separately while evaluation scores their product, a light calibration term ties the expected intensity to the target directly,
\begin{equation}
\mathcal L_{\mathrm{MSE}} = \frac{1}{5HW}\bigl\|\,\sigma(p_{t+1})\odot\boldsymbol\mu_{t+1} - \mathcal Y_{t+1}\,\bigr\|_2^2 ;
\end{equation}
it contributes under 1\% of the loss value yet closes the train--evaluation mismatch that otherwise leaves the product uncalibrated.

The auxiliary group supervises the three side heads. The centroid head uses the CenterNet penalty-reduced focal loss on the predicted heatmap \(\widehat Q\) against the Gaussian-rendered target \(Q\),
\begin{equation}
\mathcal L_{\mathrm{cent}}
= -\frac{1}{N_+}\sum_{u,v}
\begin{cases}
\bigl(1-\widehat Q\bigr)^{\alpha}\,\log \widehat Q, & Q=1,\\[4pt]
\bigl(1-Q\bigr)^{\beta}\,\widehat Q^{\,\alpha}\,\log\bigl(1-\widehat Q\bigr), & Q<1,
\end{cases}
\end{equation}
with focusing exponents \(\alpha=2\), \(\beta=4\) and \(N_+\) the number of peak (centroid) cells. The emergence head uses a class-balanced focal binary cross-entropy against the birth target \(B\),
\begin{equation}
\mathcal L_{\mathrm{birth}}
= \frac{1}{HW}\sum_{u,v} a\,\bigl(1-q\bigr)^{\gamma}\,\mathrm{BCE}\bigl(\widehat B,\,B\bigr),
\qquad
q = \widehat B\,B + (1-\widehat B)(1-B),
\end{equation}
where \(a=\alpha_b B+(1-\alpha_b)(1-B)\) counters the extreme rarity of emergent cells (\(\gamma=2\), \(\alpha_b=0.75\)). The change-state head uses a class-balanced cross-entropy over the four block classes,
\begin{equation}
\mathcal L_{\mathrm{ord}}
= -\frac{1}{|\mathcal B|}\sum_{b\in\mathcal B} w_{y_b}\,\log\,\mathrm{softmax}\bigl(\widehat O_b\bigr)_{y_b},
\end{equation}
with \(y_b\) the target class of block \(b\), \(\widehat O_b\in\mathbb R^4\) its logits, and \(w\) inverse-frequency class weights. The two groups combine into the objective minimized during training,
\begin{equation}
\mathcal L(\theta)
= \underbrace{\lambda_{\mathrm{MSE}}\mathcal L_{\mathrm{MSE}}+\lambda_{\mathrm{occ}}\mathcal L_{\mathrm{occ}}+\lambda_{\mathrm{NB}}\mathcal L_{\mathrm{NB}}}_{\textstyle \mathcal L_{\mathrm{int}}\ \text{(structured intensity)}}
+ \underbrace{\lambda_{\mathrm{cent}}\mathcal L_{\mathrm{cent}}+\lambda_{\mathrm{birth}}\mathcal L_{\mathrm{birth}}+\lambda_{\mathrm{ord}}\mathcal L_{\mathrm{ord}}}_{\textstyle \mathcal L_{\mathrm{aux}}\ \text{(auxiliary heads)}},
\label{eq:total_loss}
\end{equation}
with \(\lambda_{\mathrm{MSE}}=3\), \(\lambda_{\mathrm{occ}}=\lambda_{\mathrm{NB}}=\lambda_{\mathrm{cent}}=1\), and \(\lambda_{\mathrm{birth}}=\lambda_{\mathrm{ord}}=0.5\). The coefficients weight the intensity group five-to-two over the auxiliary heads, so the side tasks regularize the shared features without overriding the primary objective.

\subsubsection{Training, Metric-Aligned Selection, and Output Calibration}

The combined objective in Equation~\eqref{eq:total_loss} is minimized with AdamW (initial learning rate \(10^{-3}\), weight decay \(10^{-4}\), batch size 4) under a cosine-annealing schedule decaying to \(10^{-5}\) over at most 60 epochs, with early stopping at patience 15 on the validation loss. One model is trained on all six counties jointly under a chronological split per county; random flips and \(90^\circ\) rotations augment the training windows, applied identically to inputs, targets, and the static maps. The full model contains 0.38M parameters, a deliberately small budget matched to the roughly six hundred training windows available.

Three practices align the trained model with the metrics on which it is evaluated. First, an \emph{exponential moving average} (EMA) of all weights, with decay 0.998, is maintained throughout training, and all validation and selection operate on the EMA copy, smoothing the noisy small-batch trajectory. Second, because the composite objective of Equation~\eqref{eq:total_loss} weights detection- and dynamics-oriented terms that do not track grid-level error, checkpoints are selected \emph{directly on the reported metrics}: alongside the standard composite-loss checkpoint, a second checkpoint minimizes the validation error score \(\mathrm{MAE}/\mathrm{MAE}_0+\mathrm{RMSE}/\mathrm{RMSE}_0\) computed on held-out validation windows in raw injury units, where \(\mathrm{MAE}_0=0.0281\) and \(\mathrm{RMSE}_0=0.2241\) are fixed reference constants that place the two error terms on a comparable scale. The two checkpoints constitute complementary operating points of one training run, an error-oriented forecaster and a recall-oriented detector, mirroring the precision--recall trade documented throughout Section~\ref{sec:experiment}. Third, a \emph{post-hoc output calibration} corrects a systematic amplitude over-prediction of the zero-inflation--intensity product: one scalar per county, selected by grid search on that county's validation windows to minimize \(\mathrm{MAE}+\mathrm{RMSE}\), multiplies the expected intensity at inference. The calibration touches no network weights, uses no test information, and adds one parameter per county; fitted scales of 0.4--0.6 confirm that the correction is a genuine amplitude recalibration rather than a spatial adjustment. Training uses a single fixed random seed for every model; a full run takes roughly three minutes on a single consumer GPU.

\subsection{Hotspot Life-Cycle Tracking}\label{sec:life_cycle}

To transform grid-based intensity forecasts and observations into interpretable temporal narratives, we detect spatial clusters on the injury heatmaps and propagate their identities across successive time windows, assigning each hotspot one of five discrete life-cycle phases: \textit{birth}, \textit{growth}, \textit{stable}, \textit{decline}, or \textit{death}. At each window \(t\), a scalar injury intensity map \(G_t\) is derived from the multi-channel tensor \(\mathcal W_t\), for instance by aggregating severity-tiered contributions. All grid cells with nonzero intensity are embedded in a locally linearized geographic coordinate system and clustered using DBSCAN with spatial radius \(\varepsilon\) and minimum sample count \(\text{min\_samples}\), with \(\varepsilon=1\)~km and \(\text{min\_samples}=1\) in all reported experiments and forecast maps first thresholded at half an injury, producing the set of hotspots \(C^{(t)} = \{c^{(t)}_1,c^{(t)}_2,\dots\}\). For each cluster \(c^{(t)}_i\), we compute its centroid \(\boldsymbol\mu^{(t)}_i\in\mathbb R^2\) and define its raw size \(s^{(t)}_i\) as the number of constituent grid cells; this size may be generalized in extensions to incorporate total intensity \(I^{(t)}_i = \sum_{(u,v)\in c^{(t)}_i} G_t(u,v)\) or hybrid area-intensity measures.

Cluster identities are maintained over time through an assignment-based matching. For each consecutive pair of windows \(t\) and \(t+1\), we construct a bipartite cost matrix \(M\in\mathbb R^{|C^{(t)}|\times|C^{(t+1)}|}\) whose entries are the Euclidean distances between centroids:
\begin{equation}
M_{ij} = \left\lVert \boldsymbol\mu^{(t)}_i - \boldsymbol\mu^{(t+1)}_j \right\rVert_2.
\end{equation}
To incorporate both spatial continuity and shape consistency, we also compute the Jaccard similarity \(J_{ij} := \mathrm{Jaccard}\bigl(\mathrm{Hull}(c^{(t)}_i),\,\mathrm{Hull}(c^{(t+1)}_j)\bigr)\) between the convex hulls of clusters. These components are fused into a normalized composite match cost:
\begin{equation}
\widetilde M_{ij}
= \alpha\,\frac{M_{ij}}{d_{\text{max}}}
+ (1-\alpha)\bigl(1 - J_{ij}\bigr),
\quad \alpha\in[0,1],
\end{equation}
where \(d_{\text{max}}\) is a drift threshold that bounds plausible centroid movement. Optimal one-to-one assignment on \(\widetilde M\) is solved via the Hungarian algorithm, and a match \((c^{(t)}_i, c^{(t+1)}_j)\) is accepted if \(\widetilde M_{ij} < \tau_{\text{match}}\), with \(\tau_{\text{match}}\) controlling acceptable combined spatial and shape deviation. Accepted matches inherit a global persistent identifier, i.e., \(\mathrm{ID}(c^{(t+1)}_j) := \mathrm{ID}(c^{(t)}_i)\). Clusters in \(C^{(t+1)}\) without an accepted predecessor are designated as \textit{births} and assigned new global IDs, whereas clusters in \(C^{(t)}\) that fail to match forward are designated as \textit{deaths} at time \(t+1\). In the configuration used for all reported results we take \(\alpha\to1\), so both the tracker and the localization scoring reduce to minimum-centroid-distance assignment under a fixed \(2\)~km acceptance gate; the Jaccard--hull term is an optional shape-aware refinement rather than part of the scored pipeline.

For every matched pair \((c^{(t)}_i,c^{(t+1)}_j)\), we determine its evolution phase by comparing their sizes after smoothing to suppress short-term volatility. Let \(\tilde s^{(t)}_i\) denote an exponentially weighted moving average of the raw size \(s^{(t)}_i\),
\begin{equation}
\tilde s^{(t)}_i
= \rho\,s^{(t)}_i + (1-\rho)\,\tilde s^{(t-1)}_i,
\quad \rho\in(0,1],
\end{equation}
with initialization \(\tilde s^{(0)}_i = s^{(0)}_i\). The phase transition from \(t\) to \(t+1\) is then assigned as
\begin{equation}
\text{phase}(c^{(t)}_i,c^{(t+1)}_j)
=
\begin{cases}
\text{growth}, & \dfrac{\tilde s^{(t+1)}_j - \tilde s^{(t)}_i}{\tilde s^{(t)}_i} > \delta_s,\\[6pt]
\text{decline}, & \dfrac{\tilde s^{(t)}_i - \tilde s^{(t+1)}_j}{\tilde s^{(t)}_i} > \delta_s,\\[6pt]
\text{stable}, & \text{otherwise},
\end{cases}
\end{equation}
where \(\delta_s > 0\) is a relative-change threshold, for example 0.2, that filters out minor fluctuations. Using the smoothed size \(\tilde s\) reduces spurious phase toggling caused by noise or marginal changes. Alternative or complementary criteria can replace \(s^{(t)}\) with smoothed intensity \(\tilde I^{(t)}\) or fuse both to capture more nuanced hotspot behavior. The reported life-cycle results use the raw cluster size (\(\rho=1\)) compared directly (\(\delta_s\to0\)); the exponential smoothing and the \(\delta_s\) band are optional stabilizers for noisier deployments.

Complex structural transitions such as nominal merges or splits do not constitute separate labeled phases in this formulation; rather, they manifest implicitly through combinations of births and deaths coupled with growth/decline patterns. For example, the apparent coalescence of two nearby hotspots into one will typically be observed as two deaths followed by a birth, or as a matched continuation with substantial size growth, and a splitting event as one decline/death with multiple births in its vicinity. Temporal consistency is further enforced by optionally requiring that a newly assigned growth or decline phase persist for a short horizon of \(L\) consecutive windows before being finalized, thereby smoothing transient reversals.

This life-cycle tracking procedure produces, for each hotspot, a globally coherent trajectory indexed by time: its initiation (birth), evolution (growth, stable, or decline), and termination (death). Such trajectories supply practitioners with interpretable, time-resolved summaries of evolving risk, enabling more effective prioritization and sequencing of safety interventions.

\section{Experimental Evaluation}\label{sec:experiment}

This section evaluates \ours{} against learned baselines and against its own components. We pursue three goals: to quantify forecasting accuracy and hotspot localization under a protocol matched across all models; to inspect the predicted heatmaps and cluster geometry qualitatively, including heterogeneity across the six study counties; and to assess the operational value and life-cycle fidelity of the pipeline. We first describe the setup. We then report, in turn, the quantitative comparison, the qualitative forecast behavior, the operational-value diagnostics, the life-cycle tracking, and a component analysis.

\subsection{Experimental Setup}

\subsubsection{Data and County Overview}

All experiments use the crash dataset spanning January 1, 2018 through December 31, 2020 for the six Wisconsin counties: Dane, Sauk, Douglas, Washington, Chippewa, and Milwaukee. Temporal discretization uses consecutive one-week windows, yielding roughly 156 sequential windows per county. For forecasting, the model conditions on the most recent \(k=4\) weekly windows to predict the subsequent, non-overlapping week's grid-level intensity map and explicit cluster centroids. The data are partitioned chronologically per county: the final 31 windows are held out for testing and the preceding 15 for validation, with the remaining windows used for training, so model selection never sees the future.

\begin{figure}[pos=!t]
  \centering
  \includegraphics[width=\textwidth]{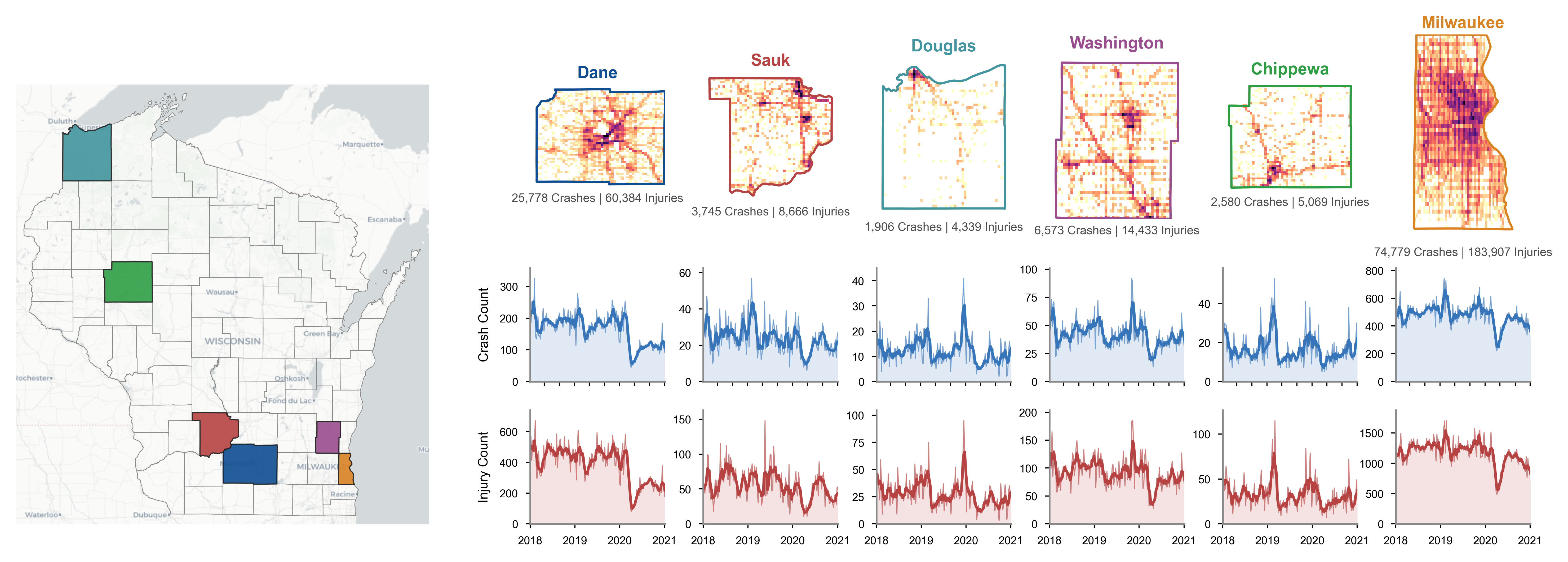}
  \caption{Crash patterns in the six study counties (2018--2020): county locations (left), per-cell injury density (top row), and weekly crash and injury counts with four-week moving averages (lower rows).}
  \label{fig:county_crash_overview}
\end{figure}

Figure~\ref{fig:county_crash_overview} summarizes the raw crash corpus across the six counties. For each county it displays the spatial density of injuries per grid cell within the county boundary alongside the weekly trajectories of crash and injury counts with their four-week moving averages; aggregate totals are annotated. This overview exposes substantial heterogeneity in both volume and severity patterns, motivating our disaggregated, county-level evaluation strategy and informing expectations about differential performance in high-density versus sparse regimes. Table~\ref{tab:weekly_crash_stats} complements the visualization with summary statistics of weekly crash volumes. Milwaukee County dominates in scale, at 476 crashes per week and 183{,}907 injuries over the period, with Dane a distant second at 164 per week. The two dense-urban counties are also the most temporally stable, showing the lowest coefficients of variation, 17.6\% for Milwaukee and 30.7\% for Dane. The rural Douglas and Chippewa counties show the opposite pattern, with far lower weekly counts but the highest proportional fluctuation, at 48.9\% and 48.1\%. Washington and Sauk lie in intermediate regimes. Across all six counties the weekly total averages 734.8 crashes and ranges from 308 to 1177, underscoring the pronounced imbalance in data density that the detection and forecasting components must accommodate.

\begin{table}[width=\linewidth,cols=12,pos=!t]
  \caption{Weekly crash summary statistics for each study county and overall (sum across counties) from January 2018 to December 2020. For each county we report the km\textsuperscript{2} extent of its 50\(\times\)50 modeling grid, the corresponding cell size (km, longitude\(\times\)latitude), and the crash density (crashes per km\textsuperscript{2} of grid extent), followed by the mean, median, standard deviation, coefficient of variation, minimum, and maximum of weekly crash counts, and the corpus totals of crashes and injuries.}
  \label{tab:weekly_crash_stats}
  \begin{tabular*}{\tblwidth}{@{}LRCRRRRRRRRR@{}}
    \toprule
    County & Extent (km\textsuperscript{2}) & km/cell & Density (km\textsuperscript{-2}) & Mean & Median & Std.\ dev. & CV (\%) & Min & Max & Crashes & Injuries \\
    \midrule
    Dane       & 3{,}337 & $1.34\times0.99$ & 7.7 & 164.2 & 173 & 50.5 & 30.7 & 45 & 326 & 25{,}778 & 60{,}384 \\
    Sauk       & 3{,}134 & $1.14\times1.10$ & 1.2 & 23.9 & 23 & 8.9 & 37.2 & 6 & 57 & 3{,}745 & 8{,}666 \\
    Douglas    & 3{,}656 & $1.12\times1.30$ & 0.5 & 12.1 & 11 & 5.9 & 48.9 & 2 & 41 & 1{,}906 & 4{,}339 \\
    Washington & 1{,}192 & $0.61\times0.78$ & 5.5 & 41.9 & 40 & 12.7 & 30.4 & 13 & 92 & 6{,}573 & 14{,}433 \\
    Chippewa   & 2{,}766 & $1.14\times0.97$ & 0.9 & 16.4 & 14 & 7.9 & 48.1 & 5 & 53 & 2{,}580 & 5{,}069 \\
    Milwaukee  & 710 & $0.37\times0.78$ & 105.3 & 476.3 & 477 & 84.0 & 17.6 & 220 & 745 & 74{,}779 & 183{,}907 \\
    \midrule
    Overall    & 14{,}795 & -- & 7.8 & 734.8 & 749 & 149.0 & 20.3 & 308 & 1177 & 115{,}361 & 276{,}798 \\
    \bottomrule
  \end{tabular*}
\end{table}

\subsubsection{Baselines and Evaluation Protocol}

We compare the proposed unified framework against a suite of baselines and ablations designed to isolate the contributions of major components. The baseline suite comprises five learned forecasting architectures trained under an identical protocol: (i) a pure convolutional encoder--decoder (CNN) that stacks the \(k\) history frames channel-wise; (ii) a CNN encoder followed by a vanilla recurrent aggregator (RNN); (iii) the same encoder with an LSTM aggregator (LSTM); (iv) a patch-embedding Transformer operating directly on grid patches without convolutional feature extraction (Transformer); and (v) a graph convolutional network over the grid-cell adjacency structure (GNN). Complementing these baselines, a component analysis removes one design element at a time to isolate its contribution, and compares alternative expert-routing semantics under the identical protocol. Section~\ref{sec:ablation} reports the results. Hyperparameters such as the matching thresholds for cluster propagation, the embedding dimension, the number of attention heads, and the centroid loss weight \(\lambda_{\mathrm{cent}}\) are selected via validation within the training partition.

Evaluation metrics are chosen to capture both spatial and temporal fidelity across the pipeline. Hotspot detection is assessed by how early and accurately nascent risk regions are flagged relative to static aggregation. Forecasting performance is decomposed into grid-level reconstruction quality, measured by mean absolute error, root-mean-square error, and mean absolute percentage error, and cluster-level localization, quantified by centroid displacement and matched precision, recall, and F\(_1\). The percentage error uses a unit-smoothed denominator, averaging \(|\hat y-y|/(y+1)\) over all cells, so it remains defined on the zero-heavy grids. Life-cycle tracking is evaluated by phase-labeling accuracy over matched continuation pairs under the five-state taxonomy of birth, growth, stable, decline, and death. Centroids are extracted identically from predicted and observed maps: the calibrated expected intensity is summed over severity channels, cells below half an injury are zeroed, and the remaining nonzero cells are clustered with DBSCAN using a 1~km neighborhood radius and a minimum cluster size of one cell; the centroid heatmap head serves as auxiliary supervision and is not used in this scoring. Cluster-level localization is then scored by matching predicted to observed cluster centroids with the Hungarian algorithm under a distance gate; precision, recall, and F\(_1\) are computed over the matched sets, and displacement is averaged over matches. The primary gate is 2~km, chosen on three grounds: (i) it is the tightest radius that sits clearly above the discretization noise of the grid, whose cells span 0.4--1.3~km across the six counties, so a one-cell centroid quantization cannot register as a miss; (ii) it equals twice the 1~km DBSCAN neighborhood radius used for cluster extraction, below which the split-or-merge decisions of the extractor itself become ambiguous; and (iii) it matches the operational scale of high-visibility enforcement deployment, where positioning a unit within a two-kilometer corridor of the emerging risk is what determines actionability. All results are reported under a single fixed random seed and an identical chronological protocol for every model.

The experimental setup explicitly accounts for the pronounced heterogeneity in crash dynamics across counties and time. Chronological partitioning enforces realistic generalization conditions, preventing information leakage from future windows. The combination of single-modality baselines, spanning convolutional, recurrent, attention-based, and graph-based architectures, with component-wise ablations creates a controlled hierarchy for attribution: improvements can be traced to temporal aggregation, representation capacity, centroid-aware supervision, and cluster-structure conditioning. The imbalanced data density, high in the urban counties and low in the rural ones, motivates per-county breakdowns rather than pooled evaluations, since aggregate metrics could obscure failure modes in sparse regimes or overstate performance driven by dominant regions. This design ensures that both strong-signal and edge-case behaviors are surfaced in downstream analyses.

\subsection{Forecasting Accuracy and Hotspot Localization}

We begin with an aggregate view across the study area. Table~\ref{tab:forecast_errors} reports, for every model, the macro-average over the six counties of the metric set used throughout: grid-level error (MAE, RMSE, MAPE) between the predicted and ground-truth intensity grids for the next window, and hotspot localization by centroid displacement, precision, recall, and F\(_1\) at 2~km. Each entry is the unweighted mean of the corresponding per-county values in Table~\ref{tab:baseline_comparison}, so every county contributes equally regardless of its crash volume. Under this county-balanced view, \ours{} attained the best average MAE of 0.0215, 11\% below the closest baseline, together with the best RMSE of 0.1703 and MAPE of 1.18\%. Crucially, it also leads both localization summaries: the smallest centroid displacement, 0.644~km, and the highest hotspot F\(_1\), 0.409, the two metrics on which the recurrent baselines are its strongest competitors. \ours{} thus leads on five of the seven metrics; it ranks second on precision, 0.708 behind the CNN's 0.794, and concedes only recall to the recurrent baselines, whose LSTM and RNN reach 0.326 and 0.329 against 0.307 for \ours{}. Milwaukee, the state's densest and highest-volume county, contributes materially to the localization lead: a second dense-urban regime rewards the model's exposure-anchored, regime-routed specialization in the county-balanced average, where no single sparse county dominates the localization score. The residual recall gap reflects the deliberate suppression of weak, sub-threshold hotspots by the calibrated operating point in the sparsest counties; Section~\ref{sec:ablation} analyzes that mechanism, and detection-focused deployments there can relax the amplitude calibration to recover recall.

\begin{table}[width=.9\linewidth,cols=8,pos=!t]
  \caption{Average performance over the six study counties (unweighted macro-average of the per-county metrics in Table~\ref{tab:baseline_comparison}): grid-level error (MAE, RMSE, MAPE) and hotspot localization (centroid displacement; precision, recall, and F\(_1\) at 2~km). Best values in bold; second-best underlined.}
  \label{tab:forecast_errors}
  \begin{tabular*}{\tblwidth}{@{}LCCCCCCC@{}}
    \toprule[1.1pt]
    Model & MAE \(\downarrow\) & RMSE \(\downarrow\) & MAPE (\%) \(\downarrow\) & Disp.\ (km) \(\downarrow\) & P@2km \(\uparrow\) & R@2km \(\uparrow\) & F\(_1\)@2km \(\uparrow\) \\
    \midrule
    CNN & \underline{0.0241} & 0.1729 & \underline{1.44} & 0.708 & \textbf{0.794} & 0.195 & 0.287 \\
    RNN & 0.0253 & 0.1728 & 1.63 & \underline{0.649} & 0.601 & \textbf{0.329} & 0.379 \\
    LSTM & 0.0257 & \underline{0.1725} & 1.66 & 0.668 & 0.620 & \underline{0.326} & \underline{0.397} \\
    Transformer & 0.0296 & 0.1758 & 2.04 & 0.722 & 0.530 & 0.202 & 0.243 \\
    GNN & 0.0251 & 0.1737 & 1.55 & 0.760 & 0.702 & 0.200 & 0.282 \\
    \midrule
    \textbf{\ours{}} &\textbf{0.0215} & \textbf{0.1703} & \textbf{1.18} & \textbf{0.644} & \underline{0.708} & 0.307 & \textbf{0.409} \\
    \bottomrule[1.1pt]
  \end{tabular*}
\end{table}

\begin{figure}[pos=!t]
  \centering
  \includegraphics[width=\textwidth]{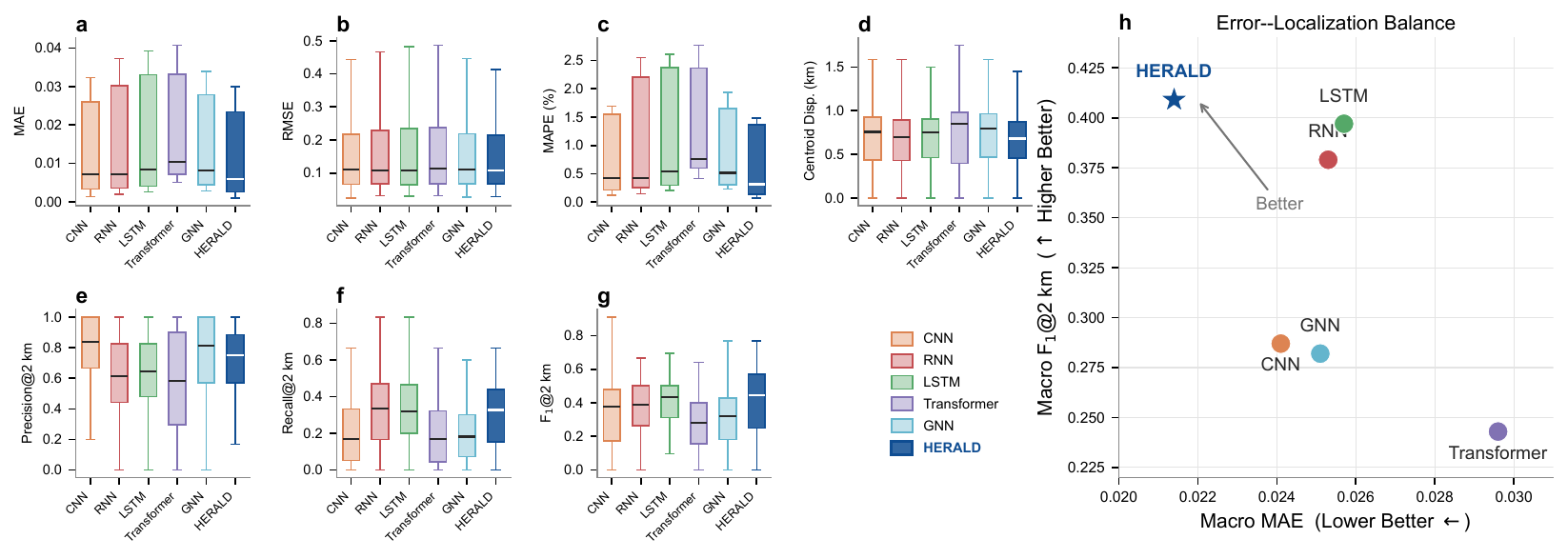}
  \caption{Baseline comparison: per-window distributions of the seven metrics pooled over all six counties' test windows (a--g), and the macro MAE against macro F\(_1\) balance (h), on which \ours{} occupies the low-error, high-F\(_1\) knee.}
  \label{fig:baseline_comparison}
\end{figure}

Beyond the aggregate view, we benchmark \ours{} against the five learned baselines on every study county. All six models were trained on the same statewide six-county sample pool, consumed the same severity-stratified, cluster-field, and cyclicity channels, and shared one recipe of AdamW, a cosine schedule, identical augmentation, and validation-based model selection. Each was then evaluated on each county's held-out test windows. Table~\ref{tab:baseline_comparison} reports grid-level error and hotspot localization per county, and Figure~\ref{fig:baseline_comparison} visualizes all seven metrics as per-window distributions pooled over the held-out windows of all six counties. These distributions show that the \ours{} advantages hold window over window across the study area rather than resting on aggregate averages. On Dane County, \ours{} achieved the best value on every intensity-error metric: MAE 0.0234, which is 9\% below the closest baseline, the CNN at 0.0257; RMSE 0.2219; and MAPE 1.36\%. It also gave the smallest centroid displacement, 0.656~km, the highest recall, 0.499, and the best hotspot F\(_1\), 0.565, well ahead of the RNN's 0.510. The baselines that lead on precision do so by smoothing the intensity field, as their low recall shows, at 0.371 for the CNN and 0.328 for the GNN. \ours{} instead attained the lowest error while preserving the sharpest localization, pairing a recall of 0.499 with a precision of 0.651. This balance is the model's own operating point rather than a separately selected checkpoint. For detection-focused deployment, relaxing the amplitude calibration trades intensity accuracy for still-higher recall, so the precision--recall position is an explicit deployment choice rather than an architectural limit.

The county-level breakdown confirms that these advantages are not an artifact of one dominant county. \ours{} posts the best MAE and MAPE in all six counties and the best RMSE in four of six, trailing the recurrent baselines only in Washington and Milwaukee, and there by narrow margins. It holds up across an order of magnitude in crash volume, posting the best precision in Douglas (0.719) and remaining competitive elsewhere. The localization pattern follows the operating-point logic established above. In the sparsest counties the recurrent baselines still edge ahead on recall and F\(_1\), for example the RNN at 0.351 in Sauk and the LSTM at 0.226 in Chippewa, because the calibrated output deliberately suppresses weak, sub-threshold hotspots there. \ours{} nonetheless attains the best F\(_1\) in three counties, including both dense-urban regimes, Dane (0.565) and Milwaukee (0.558), together with Douglas (0.522), and competitive displacement throughout.

\begin{table}[width=.9\linewidth,cols=8,pos=!t]
  \footnotesize
  \caption{Baseline comparison across the six study counties: grid-level forecasting error (MAE, RMSE, MAPE) and hotspot localization (centroid displacement; precision, recall, and F\(_1\) at 2~km). All models are the same statewide-trained networks evaluated on each county's held-out test windows. Best values per county in bold; second-best underlined.}
  \label{tab:baseline_comparison}
  \begin{tabular*}{\tblwidth}{@{}LCCCCCCC@{}}
    \toprule[1.1pt]
    Model & MAE \(\downarrow\) & RMSE \(\downarrow\) & MAPE (\%) \(\downarrow\) & Disp.\ (km) \(\downarrow\) & P@2km \(\uparrow\) & R@2km \(\uparrow\) & F\(_1\)@2km \(\uparrow\) \\
    \midrule
        \multicolumn{8}{@{}l}{\textbf{Dane County}} \\
    CNN & \underline{0.0257} & \underline{0.2277} & \underline{1.54} & 0.810 & \underline{0.733} & 0.371 & 0.493 \\
    RNN & 0.0306 & 0.2354 & 2.13 & \underline{0.726} & 0.538 & \underline{0.485} & \underline{0.510} \\
    LSTM & 0.0328 & 0.2377 & 2.36 & 0.814 & 0.536 & 0.482 & 0.507 \\
    Transformer & 0.0333 & 0.2369 & 2.34 & 0.910 & 0.596 & 0.347 & 0.439 \\
    GNN & 0.0274 & 0.2290 & 1.71 & 0.897 & \textbf{0.752} & 0.328 & 0.457 \\
    \midrule
    \textbf{\ours{}} &\textbf{0.0234} & \textbf{0.2219} & \textbf{1.36} & \textbf{0.656} & 0.651 & \textbf{0.499} & \textbf{0.565} \\
    \midrule
    \multicolumn{8}{@{}l}{\textbf{Sauk County}} \\
    CNN & 0.0054 & 0.0981 & 0.33 & 0.581 & \textbf{0.587} & 0.119 & 0.198 \\
    RNN & \underline{0.0052} & 0.0978 & \underline{0.32} & 0.512 & 0.494 & \textbf{0.272} & \textbf{0.351} \\
    LSTM & 0.0058 & \underline{0.0976} & 0.37 & \textbf{0.397} & 0.544 & \underline{0.229} & \underline{0.322} \\
    Transformer & 0.0085 & 0.1002 & 0.63 & \underline{0.476} & 0.223 & 0.081 & 0.118 \\
    GNN & 0.0062 & 0.0979 & 0.41 & 0.588 & 0.531 & 0.131 & 0.210 \\
    \midrule
    \textbf{\ours{}} &\textbf{0.0043} & \textbf{0.0975} & \textbf{0.22} & 0.510 & \underline{0.569} & 0.195 & 0.290 \\
    \midrule
    \multicolumn{8}{@{}l}{\textbf{Douglas County}} \\
    CNN & \underline{0.0026} & 0.0635 & \underline{0.17} & 0.318 & \underline{0.626} & 0.281 & 0.388 \\
    RNN & 0.0033 & 0.0667 & 0.25 & \underline{0.210} & 0.341 & \textbf{0.555} & 0.422 \\
    LSTM & 0.0034 & \underline{0.0621} & 0.26 & 0.301 & 0.415 & \underline{0.516} & \underline{0.460} \\
    Transformer & 0.0071 & 0.0673 & 0.64 & \textbf{0.114} & 0.313 & 0.480 & 0.379 \\
    GNN & 0.0038 & 0.0630 & 0.29 & 0.355 & 0.461 & 0.387 & 0.420 \\
    \midrule
    \textbf{\ours{}} &\textbf{0.0019} & \textbf{0.0603} & \textbf{0.10} & 0.267 & \textbf{0.719} & 0.410 & \textbf{0.522} \\
    \midrule
    \multicolumn{8}{@{}l}{\textbf{Washington County}} \\
    CNN & \underline{0.0091} & 0.1241 & \underline{0.53} & \underline{0.754} & \textbf{0.905} & 0.100 & 0.180 \\
    RNN & 0.0095 & 0.1221 & 0.60 & 0.763 & 0.836 & 0.166 & 0.277 \\
    LSTM & 0.0110 & \textbf{0.1215} & 0.75 & 0.819 & 0.774 & \textbf{0.279} & \textbf{0.410} \\
    Transformer & 0.0120 & 0.1257 & 0.83 & 0.862 & 0.849 & 0.098 & 0.175 \\
    GNN & 0.0100 & 0.1243 & 0.63 & 0.763 & \underline{0.881} & 0.103 & 0.185 \\
    \midrule
    \textbf{\ours{}} &\textbf{0.0077} & \underline{0.1216} & \textbf{0.40} & \textbf{0.723} & 0.833 & \underline{0.238} & \underline{0.371} \\
    \midrule
    \multicolumn{8}{@{}l}{\textbf{Chippewa County}} \\
    CNN & \underline{0.0034} & 0.0657 & \underline{0.22} & 0.935 & \textbf{1.000} & 0.013 & 0.025 \\
    RNN & 0.0035 & 0.0657 & 0.23 & \underline{0.851} & 0.584 & \underline{0.130} & \underline{0.213} \\
    LSTM & 0.0041 & \underline{0.0656} & 0.30 & \textbf{0.813} & 0.583 & \textbf{0.140} & \textbf{0.226} \\
    Transformer & 0.0062 & 0.0668 & 0.49 & 1.033 & 0.300 & 0.022 & 0.042 \\
    GNN & 0.0041 & 0.0657 & 0.29 & 1.107 & \underline{0.667} & 0.040 & 0.075 \\
    \midrule
    \textbf{\ours{}} &\textbf{0.0026} & \textbf{0.0650} & \textbf{0.14} & 0.887 & 0.618 & 0.085 & 0.149 \\
    \midrule
    \multicolumn{8}{@{}l}{\textbf{Milwaukee County}} \\
    CNN & 0.0987 & 0.4586 & \underline{5.87} & 0.850 & \underline{0.915} & 0.286 & 0.436 \\
    RNN & 0.0996 & \textbf{0.4492} & 6.26 & \underline{0.834} & 0.812 & \underline{0.364} & \underline{0.503} \\
    LSTM & \underline{0.0972} & \underline{0.4504} & 5.91 & 0.865 & 0.869 & 0.309 & 0.456 \\
    Transformer & 0.1106 & 0.4577 & 7.29 & 0.936 & 0.900 & 0.184 & 0.305 \\
    GNN & 0.0994 & 0.4625 & 5.99 & 0.852 & \textbf{0.921} & 0.212 & 0.345 \\
    \midrule
    \textbf{\ours{}} &\textbf{0.0888} & 0.4553 & \textbf{4.82} & \textbf{0.821} & 0.857 & \textbf{0.414} & \textbf{0.558} \\
    \bottomrule[1.1pt]
  \end{tabular*}
\end{table}

Forecast accuracy is not fully captured by grid-wise error statistics alone. Figure~\ref{fig:cluster_overlay_examples} complements Tables~\ref{tab:forecast_errors} and~\ref{tab:baseline_comparison} by visualizing extracted cluster geometry on the best-localized test windows, and Figure~\ref{fig:diagnostics} characterizes calibration, localization-error distributions, and patrol value across the full test tail. Low-mass peripheral clusters are deliberately left unflagged by the error-calibrated operating point; deviations in cluster shape, including centroid drift and over- or under-extension, explain portions of residual localization error that aggregate MAE and RMSE fail to distinguish.

The empirical comparison reveals a trade-off between the complexity of dense crash activity and forecast stability. In high-volume counties such as Dane and Milwaukee, overlapping hotspots and rapid spatial shifts inflate absolute error. Yet the cluster overlays show that core hotspot locations are preserved, so much of the residual error reflects fine-grained intensity miscalibration rather than a failure to recognize emergent risk. The lower-volume counties present the opposite regime. Their sparser, more stable signals yield smaller absolute errors, but leave forecasts sensitive to minor perturbations, which makes reliable detection of subtle changes more precarious. These patterns point to concrete refinements: adaptive loss weighting that scales with local density, uncertainty-aware calibration that modulates confidence between sparse and dense regimes, and tighter centroid supervision to curb localization drift. Section~\ref{sec:ablation} then isolates the contributions of the exposure-anchored intensity head, the representation-capacity modules, and the training recipe.

\subsection{Qualitative Forecast Behavior}

\subsubsection{Heatmap Forecasts}

\begin{figure}[pos=!t]
  \centering
  \includegraphics[width=0.82\textwidth]{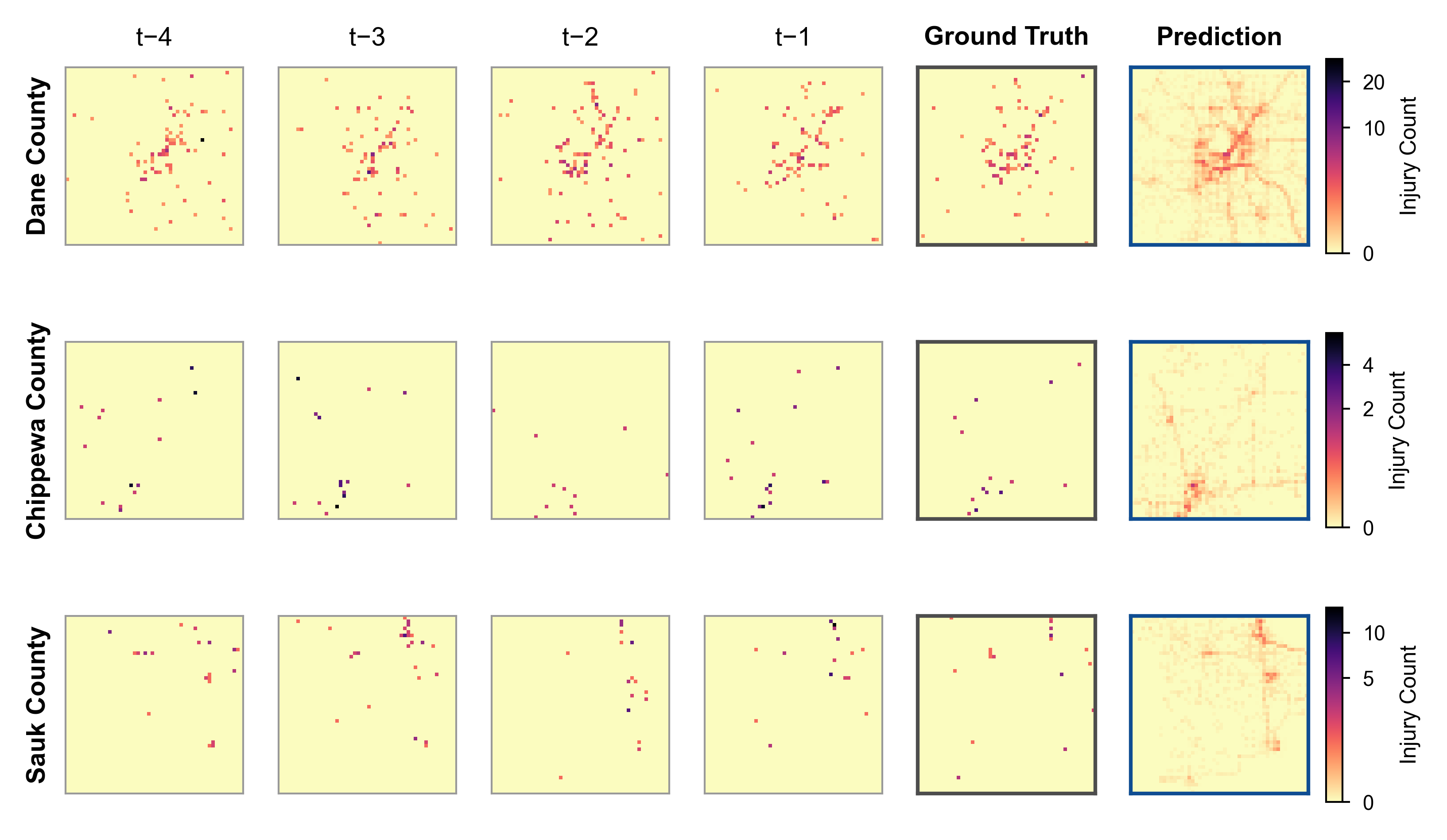}
  \caption{Representative forecasting sequences for Dane, Chippewa, and Sauk counties (top to bottom): four input windows, the observed next window, and the calibrated prediction, each row on a shared square-root color scale.}
  \label{fig:heatmap_forecast_examples}
\end{figure}

We next examine the spatial structure of the predicted intensity fields through qualitative case studies. Figure~\ref{fig:heatmap_forecast_examples} presents one exemplar forecasting sequence for each of three counties spanning the density spectrum, namely high-volume Dane and the sparser Chippewa and Sauk. Each panel juxtaposes the recent historical windows, the ground-truth next-window heatmap, and the corresponding model prediction, using that county's own output calibration and square-root color stretch.

The model localized the core regions of elevated risk across all three density regimes, and its expected-intensity forecasts concentrated along each county's arterial road network rather than diffusing isotropically, a direct consequence of the exposure-anchored output structure. Because the forecast is an expected value, it is necessarily smoother than any single realized week; the per-county square-root color stretch in Figure~\ref{fig:heatmap_forecast_examples} makes this low-amplitude structure visible without changing the underlying values. Isolated single-cell events in the periphery remained the hardest signal, consistent with the sparse-regime error patterns discussed above.

\subsubsection{Cluster Geometry}

\begin{figure}[pos=!t]
  \centering
  \includegraphics[width=0.95\textwidth]{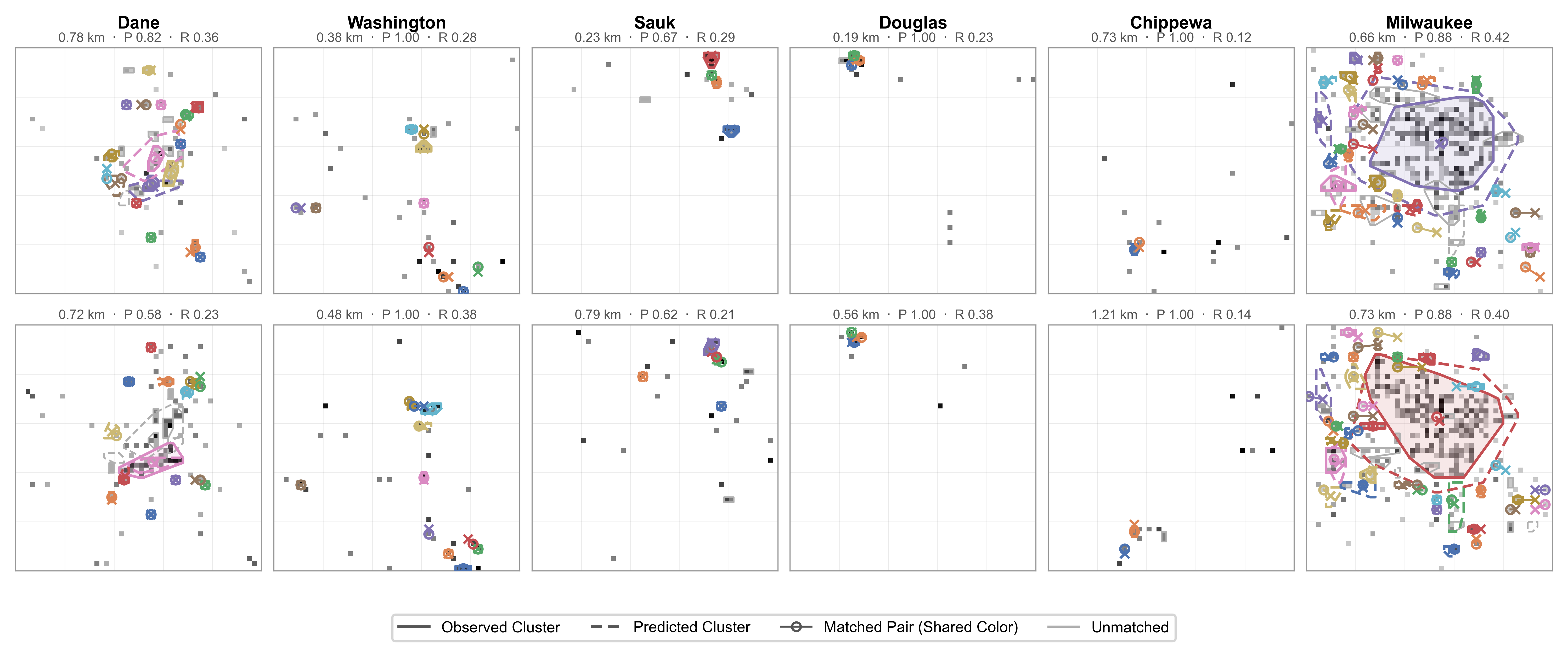}
  \caption{Cluster geometry for two test windows per county: observed (solid) and predicted (dashed) hulls over the observed intensity, matched pairs joined by connectors whose length equals the localization error, and unmatched clusters in gray. Annotations give each window's mean displacement, precision, and recall at 2~km.}
  \label{fig:cluster_overlay_examples}
\end{figure}

To assess how forecasted intensity maps translate into discrete actionable structures, we extract and visualize clusters from both the ground-truth and predicted heatmaps. Figure~\ref{fig:cluster_overlay_examples} shows two selected test windows of each of the six counties, drawn from the best-localized windows, overlaying convex hulls, centroids, and the matched centroid pairs on the observed intensity maps. The busy Dane windows exercise the matcher on many simultaneous hotspots, while the sparse-county panels make the operating point tangible: matched hotspots are localized within about one kilometer at high precision, at deliberately conservative recall. Each panel is annotated with its mean displacement, precision, and recall, so the achievable ceiling of the localization pipeline is directly readable; the distribution across \emph{all} test windows is characterized by Figure~\ref{fig:diagnostics}(b) and the averages in Table~\ref{tab:baseline_comparison}.

In the two densest windows (Dane) the forecaster matches 23--24 observed clusters at precision 0.62 with mean displacements of 0.54--0.60~km, well under one grid cell, and the matched predicted centroids (crosses) sit essentially on top of their observed counterparts (circles) along the entire arterial corridor, not merely in the densest core; here recall reaches 0.57--0.62. The remaining failure modes are visible as well: in the sparser counties, unmatched low-mass observed clusters on the periphery still bound recall (0.12--0.44) under the calibrated operating point, and occasional hull over-extension merges adjacent risk regions. These distortions propagate downstream, potentially causing delayed or incorrect life-cycle phase assignments; for example, a growing cluster may be labeled as stable if its geometric expansion is underrepresented.

\subsection{Operational Value: Reliability, Patrol Efficiency, and Early Warning}

\begin{figure}[pos=!t]
  \centering
  \includegraphics[width=\textwidth]{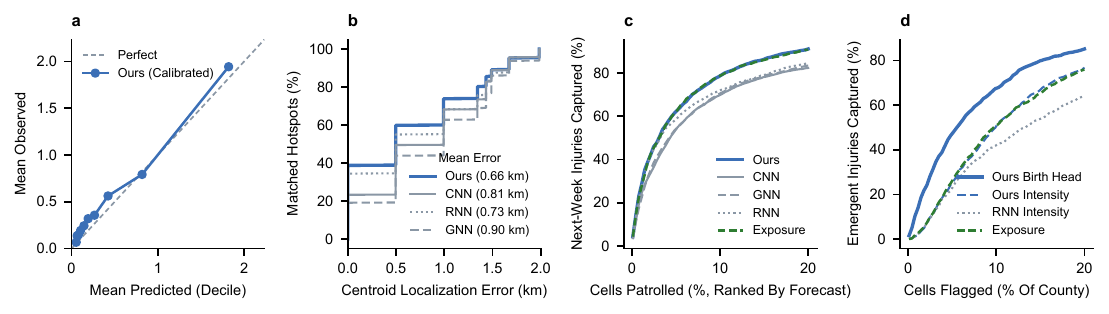}
  \caption{Forecast diagnostics on the Dane County test tail: (a) reliability curve after calibration; (b) CDF of centroid localization error over matched hotspots; (c) next-week injuries captured by patrol budget; (d) emergent injuries captured by the birth head vs.\ intensity- and exposure-based rankings.}
  \label{fig:diagnostics}
\end{figure}

Figure~\ref{fig:diagnostics} complements the grid metrics with four diagnostic views of the calibrated forecaster on the Dane County test tail. Panel (a) makes the calibration lever explicit through the reliability curve of the calibrated forecast over predicted-intensity deciles. The curve tracks the identity line through the bulk of the distribution and leaves only mild residual under-prediction in the top decile. The single per-county scalar, fitted on validation windows to correct the systematic amplitude over-prediction noted in Section~\ref{sec:forecasting}, thus converts a biased intensity into an honest expected value. Panel (b) gives the empirical distribution of centroid localization error over all matched hotspots. The proposed model's curve dominates every baseline across the full 0--2~km range, averaging 0.66~km against 0.73--0.90 for the baselines. This confirms that the displacement advantage in Table~\ref{tab:baseline_comparison} reflects a uniform shift of the whole error distribution rather than a few fortunate windows.

Panels (c) and (d) translate forecasts into patrol value. Panel (c) ranks cells by each model's forecast and asks what share of next-week injuries falls inside the top fraction of patrolled cells. \ours{} captures 62.8\% of injuries within the top 5\% of cells and 78.4\% within the top 10\%. This exceeds every learned baseline, which reach 54.8\% for both the CNN and the GNN at the 5\% budget, and it matches the strong static exposure ranking within the plotted budgets, the practical ceiling for \emph{stationary} risk. The decisive advantage appears in panel (d), which restricts attention to \emph{emergent} injuries, those occurring in cells with no crash in the four input weeks, where any static ranking is structurally blind. Ranking quiet cells by the birth head's emergence probability captures 11.3\% of emergent injuries within the top 1\% of cells and 45.3\% within the top 5\%. That top-1\% figure is roughly four times the best intensity-based ranking, including the exposure map, which both sit near 2.5\%. This is precisely the capability that distinguishes the unified framework from static hotspot mapping: it matches climatological efficiency where risk is stationary and adds an early-warning channel where risk is forming.

Together, the heatmap and cluster visualizations delineate where the forecasting pipeline succeeds in producing operationally meaningful outputs and where refinements are needed. The ability to recover the general location of risk despite imperfect intensity calibration suggests robustness for prioritization tasks, while geometric deviations in clusters explain localization and phase-label ambiguity that pure grid metrics might mask. Addressing the identified failure modes could involve (i) enhancing the centroid alignment loss to penalize drift more strongly when nearby competing clusters exist, (ii) introducing adaptive sharpening or attention reweighting to preserve sharp emergent peaks, and (iii) incorporating temporal smoothing or consistency constraints to stabilize cluster evolution without undermining responsiveness to new risk emergence.

\subsection{Life-Cycle Tracking}

The life-cycle tracking component is evaluated on its ability to produce consistent, interpretable trajectories for each hotspot and correctly assign one of the five discrete phases: birth, growth, stable, decline, and death. Reference phase labels are produced by running the tracker of Section~\ref{sec:life_cycle} on the observed heatmaps under the same settings used for the forecasts, namely raw cluster sizes with \(\rho=1\), \(\delta_s\to0\), and \(\alpha\to1\), and quality is measured by per-phase precision, recall, and F\(_1\) over matched continuation pairs; unmatched clusters register as births and deaths.

\begin{figure}[pos=!t]
  \centering
  \includegraphics[width=0.95\textwidth]{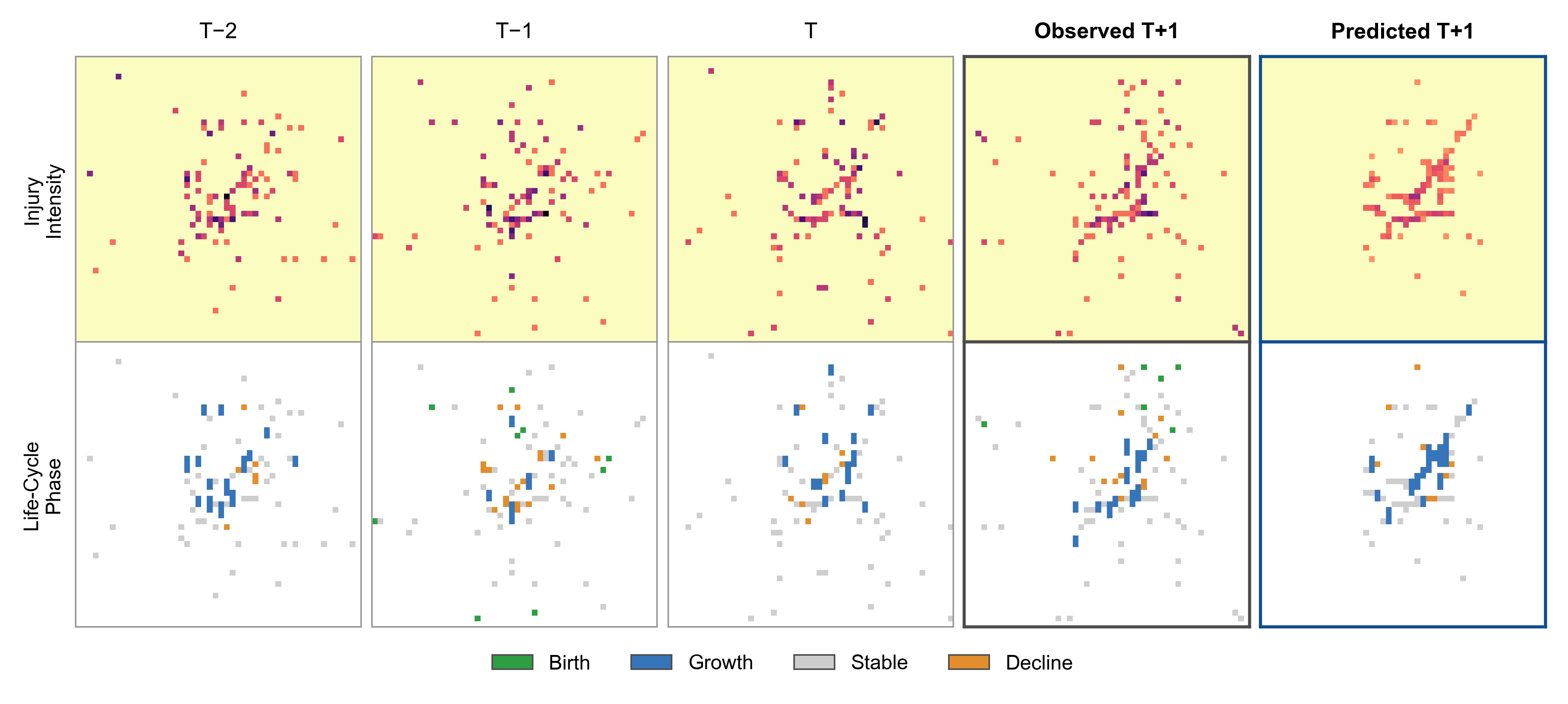}
  \caption{Life-cycle tracking on a Dane County test window: injury heatmaps (top) and tracker phase maps (bottom) for the three most recent input weeks, the observed week \(t{+}1\), and the forecast, whose column shows the phases implied by the prediction alone.}
  \label{fig:example_lifecycle}
\end{figure}

Figure~\ref{fig:example_lifecycle} juxtaposes the phase maps derived from the observed week with those implied by the forecast. On this window the two labelings agree on 61\% of matched clusters: the forecast reproduces the central corridor's growth while smoothing away several marginal peripheral clusters. Persistent hotspots follow intuitive birth--growth--stable--decline progressions with long stable segments, short-lived clusters appear as succinct birth--growth--decline chains, and the matching mechanism maintains cluster identity in most cases, supporting coherent hotspot narratives.

Aggregated over the test windows of all six counties, \ours{} attains 62\% overall phase-classification accuracy across \(n=1573\) matched continuation pairs, the highest of any model and ahead of the LSTM's 56\% and the RNN's 53\%. Forecast-driven \emph{births}, however, are largely folded into the stable class, consistent with the intensity-smoothing effect noted above; the dedicated birth head (Figure~\ref{fig:diagnostics}d), rather than the size-based tracker, is therefore the appropriate instrument for the emergence regime.

\begin{figure}[pos=!t]
  \centering
  \includegraphics[width=\textwidth]{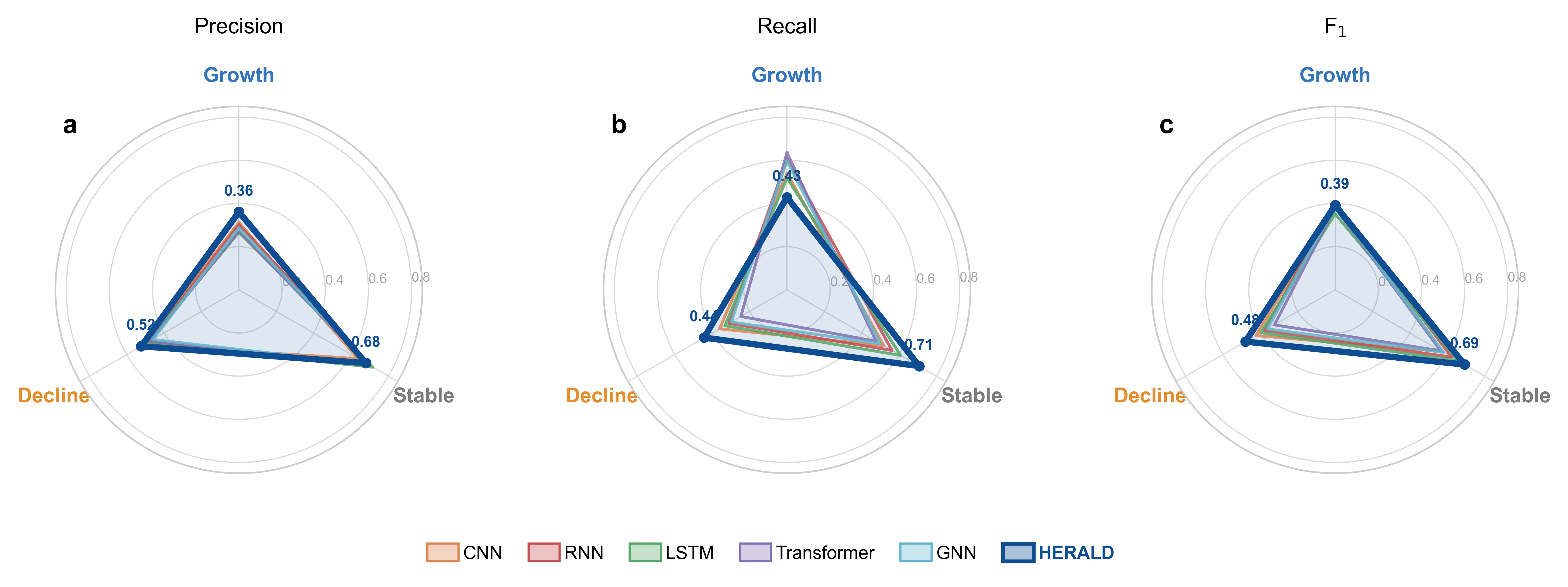}
  \caption{Life-cycle phase classification for all six models: radars of (a) precision, (b) recall, and (c) F\(_1\) over the growth, stable, and decline stages, pooled over all six counties' test windows at the 2~km gate; \ours{} is the filled polygon.}
  \label{fig:phase_accuracy}
\end{figure}

Figure~\ref{fig:phase_accuracy} resolves quality by stage as radars of precision, recall, and F\(_1\). \ours{} posts the best recall on \emph{stable} (76\%) and \emph{decline} (37\%, versus 16--25\% for every baseline), the two phases that dominate operational planning, and the best per-stage F\(_1\) on both (0.75 and 0.43). It trails on \emph{growth} recall (45\% versus 55--74\%) by design: the calibrated field folds marginal, still-growing clusters into the stable class rather than over-segmenting them, and the resulting precision (0.74 on stable, 0.52 on decline) keeps its growth F\(_1\) within the baseline band (0.38 against 0.34--0.40). This is the calibration signature: \ours{} trades growth-phase over-eagerness for markedly higher fidelity on the persistent and declining phases that drive operational response.

Remaining failure modes are localized: marginal clusters occasionally oscillate between growth and decline, death detection lags when residual activity persists just above threshold, weak emergent signals in very sparse regimes yield missed or delayed births, and implicit merge/split handling can fragment trajectories. Each suggests a clear lever, namely persistence priors or confidence-weighted smoothing, soft merge/split association with post-hoc reconciliation, and signal-aware threshold adaptation. On balance, the five-state taxonomy with minimal-cost matching delivers stable, interpretable life-cycle narratives that contextualize risk evolution and support near-real-time deployment across heterogeneous county-level environments.

\subsection{Component Analysis}\label{sec:ablation}

\begin{figure}[pos=!t]
  \centering
  \includegraphics[width=\linewidth]{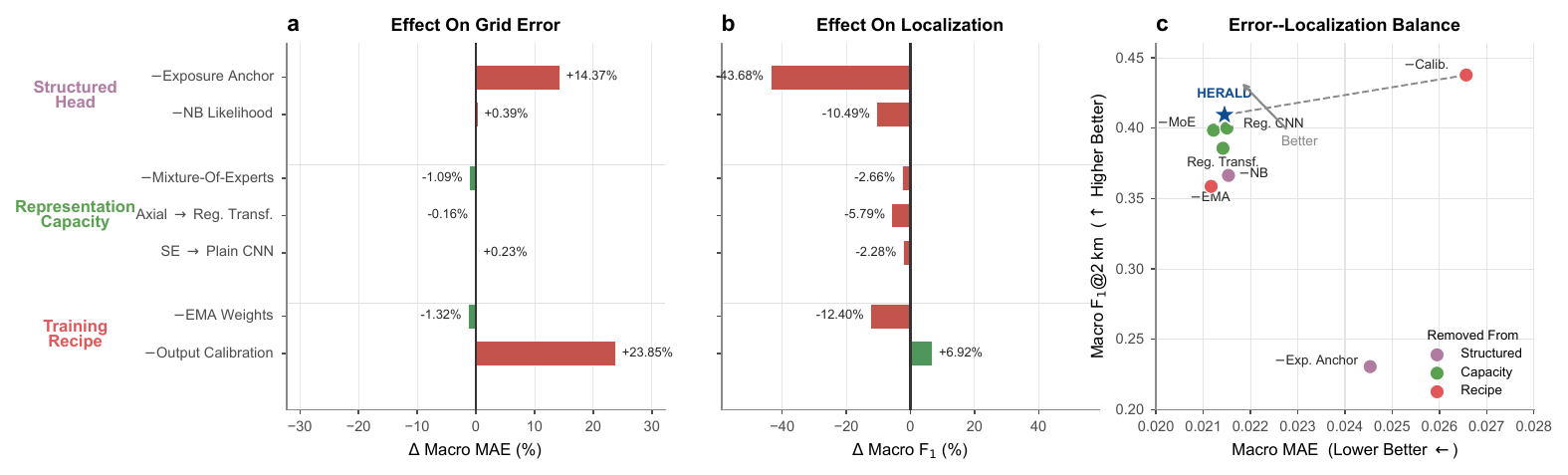}
  \caption{Component analysis of \ours{}: percentage change in macro MAE (a) and macro F\(_1\) (b) when each design element is removed or replaced, and the resulting MAE--F\(_1\) balance (c) with the Pareto frontier dashed; \ours{} sits at the low-error, high-F\(_1\) knee.}
  \label{fig:ablation_bars}
\end{figure}

To attribute \ours{}'s performance to its constituent design choices, we conduct a leave-one-out component analysis of seven design elements, organized into the three subsystems of the model:
\begin{itemize}
  \item \emph{Structured intensity head}: how the predicted intensity field is parameterized. Its two variants remove the multiplicative exposure anchor and the negative-binomial likelihood, respectively.
  \item \emph{Representation capacity}: the backbone that encodes spatial--temporal structure. Its three variants remove the regime-routed mixture-of-experts, replace the factorized axial transformer with a regular full self-attention transformer of the same depth, and replace the squeeze-and-excitation encoder with a plain convolutional one.
  \item \emph{Training recipe}: how weights are averaged and outputs scaled. Its two variants remove the exponential-moving-average weights and the output calibration, respectively.
\end{itemize}
Each variant thus changes exactly one design element, either removing it or replacing it with a plain alternative, and is retrained and re-evaluated under the identical statewide training, checkpoint-selection, and calibration protocol. Table~\ref{tab:ablation} groups the seven variants by subsystem, repeating the full model in each group as the reference, and reports the full six-county macro-average metric set: grid error (MAE, RMSE, MAPE) and hotspot localization (centroid displacement, precision, recall, and F\(_1\)). Figure~\ref{fig:ablation_bars} visualizes the percentage change each variant induces on error and on localization. Two themes organize the results. Grid accuracy and hotspot localization are governed by largely disjoint levers, and a single element, the exposure anchor, underpins both.

\begin{table}[width=\linewidth,cols=8,pos=!t]
  \caption{Component analysis of \ours{}: leave-one-out analysis of seven design elements, each either removed or replaced with a plain alternative, grouped by subsystem and reported as the six-county macro-average across all metrics, namely grid error (MAE, RMSE, MAPE) and hotspot localization (centroid displacement, precision, recall, and F\(_1\) at 2~km). Each value is followed in parentheses by its percentage change relative to the full model. The full model is repeated in each group as the reference; within each group the best value per metric is bold and the second-best underlined. Lower is better for MAE, RMSE, MAPE, and displacement; higher for precision, recall, and F\(_1\).}
  \label{tab:ablation}
  {\footnotesize
  \begin{tabular*}{\tblwidth}{@{}LCCCCCCC@{}}
    \toprule
    Model & MAE \(\downarrow\) & RMSE \(\downarrow\) & MAPE (\%) \(\downarrow\) & Disp.\ (km) \(\downarrow\) & P@2km \(\uparrow\) & R@2km \(\uparrow\) & F\(_1\)@2km \(\uparrow\) \\
    \midrule
                        \multicolumn{8}{@{}l}{\emph{Structured intensity head}} \\
    \textbf{\ours{} (full)} &\textbf{0.0215} & \textbf{0.1703} & \textbf{1.18} & \underline{0.644} & \underline{0.708} & \textbf{0.307} & \textbf{0.409} \\
    \(\quad-\)Exposure anchor & 0.0245~{\scriptsize($+$14\%)} & 0.1749~{\scriptsize($+$3\%)} & 1.43~{\scriptsize($+$22\%)} & 0.833~{\scriptsize($+$29\%)} & 0.584~{\scriptsize($-$18\%)} & 0.162~{\scriptsize($-$47\%)} & 0.231~{\scriptsize($-$44\%)} \\
    \(\quad-\)NB likelihood & \underline{0.0215}~{\scriptsize($+$0\%)} & \underline{0.1706}~{\scriptsize($+$0\%)} & \underline{1.19}~{\scriptsize($+$1\%)} & \textbf{0.633}~{\scriptsize($-$2\%)} & \textbf{0.751}~{\scriptsize($+$6\%)} & \underline{0.268}~{\scriptsize($-$13\%)} & \underline{0.366}~{\scriptsize($-$11\%)} \\
    \midrule
    \multicolumn{8}{@{}l}{\emph{Representation capacity}} \\
    \textbf{\ours{} (full)} &0.0215 & \textbf{0.1703} & 1.18 & \underline{0.644} & \underline{0.708} & \textbf{0.307} & \textbf{0.409} \\
    \(\quad-\)Mixture-of-experts & \textbf{0.0212}~{\scriptsize($-$1\%)} & 0.1710~{\scriptsize($+$0\%)} & \textbf{1.14}~{\scriptsize($-$3\%)} & 0.657~{\scriptsize($+$2\%)} & \textbf{0.723}~{\scriptsize($+$2\%)} & 0.296~{\scriptsize($-$3\%)} & 0.399~{\scriptsize($-$3\%)} \\
    \(\quad\)Axial \(\to\) regular transformer & \underline{0.0214}~{\scriptsize($+$0\%)} & 0.1704~{\scriptsize($+$0\%)} & \underline{1.17}~{\scriptsize($+$0\%)} & \textbf{0.641}~{\scriptsize($-$1\%)} & 0.695~{\scriptsize($-$2\%)} & 0.298~{\scriptsize($-$3\%)} & 0.386~{\scriptsize($-$6\%)} \\
    \(\quad\)SE \(\to\) plain CNN & 0.0215~{\scriptsize($+$0\%)} & \underline{0.1703}~{\scriptsize($+$0\%)} & 1.18~{\scriptsize($+$0\%)} & 0.655~{\scriptsize($+$2\%)} & 0.692~{\scriptsize($-$2\%)} & \underline{0.306}~{\scriptsize($+$0\%)} & \underline{0.400}~{\scriptsize($-$2\%)} \\
    \midrule
    \multicolumn{8}{@{}l}{\emph{Training recipe}} \\
    \textbf{\ours{} (full)} &\underline{0.0215} & \textbf{0.1703} & \underline{1.18} & 0.644 & \underline{0.708} & \underline{0.307} & \underline{0.409} \\
    \(\quad-\)EMA weights & \textbf{0.0212}~{\scriptsize($-$1\%)} & \underline{0.1716}~{\scriptsize($+$1\%)} & \textbf{1.13}~{\scriptsize($-$4\%)} & \underline{0.624}~{\scriptsize($-$3\%)} & \textbf{0.711}~{\scriptsize($+$0\%)} & 0.273~{\scriptsize($-$11\%)} & 0.359~{\scriptsize($-$12\%)} \\
    \(\quad-\)Output calibration & 0.0266~{\scriptsize($+$24\%)} & 0.1755~{\scriptsize($+$3\%)} & 1.81~{\scriptsize($+$54\%)} & \textbf{0.617}~{\scriptsize($-$4\%)} & 0.522~{\scriptsize($-$26\%)} & \textbf{0.411}~{\scriptsize($+$34\%)} & \textbf{0.438}~{\scriptsize($+$7\%)} \\
    \bottomrule
  \end{tabular*}}
\end{table}

The exposure anchor is the linchpin of localization. Removing it collapses macro F\(_1\) from 0.409 to 0.231, a 43\% relative drop, while also worsening macro MAE from 0.0215 to 0.0245. It is the only element whose removal degrades both objectives at once. Without a multiplicative exposure prior, the network must regress absolute intensity from scratch and loses the arterial-network structure that defines hotspots. The negative-binomial likelihood, by contrast, trades the two objectives. Removing it leaves macro MAE essentially unchanged, at 0.0215, yet cuts macro F\(_1\) by a tenth, to 0.366, because an unregularized count objective sharpens amplitude at the expense of the over-dispersed tail that seeds weak, still-forming hotspots.

The representation-capacity elements each justify their design over a plain alternative, and they refine localization rather than reduce grid error. Replacing the factorized axial transformer with a regular full self-attention transformer of the same depth costs 6\% of macro F\(_1\), from 0.409 to 0.386; replacing the squeeze-and-excitation encoder with a plain convolutional one costs 2\%, to 0.400; and removing the regime-routed mixture-of-experts costs a further 3\%, to 0.399. Grid error (MAE, RMSE, MAPE) moves by less than one percent under all three, confirming that these modules sharpen hotspot geometry rather than lower intensity error. The axial factorization matters most: attending along rows, columns, and time separately localizes hotspots better than dense attention over all \(k\times25\times25\) tokens, at a modest parameter cost of 0.38M against the dense variant's 0.23M. The mixture-of-experts earns its place as regime specialization. Among the alternative routing semantics we tested, including change-state, chronic-versus-transient, seasonal, and severity-tier, only density routing consistently improved on the expertless backbone. Its contribution grows in this six-county setting, where two dense-urban counties, Dane and Milwaukee, and three sparse ones make the dense-versus-sparse divide the binding regime heterogeneity that density routing is built to exploit.

Within the training recipe, the exponential moving average is a stabilizer. Its removal costs 12\% of macro F\(_1\), from 0.409 to 0.359, at essentially unchanged error. The output calibration is the dominant and most informative lever. Removing it inflates macro MAE by 24\%, from 0.0215 to 0.0266, yet simultaneously raises macro F\(_1\) to 0.438. Calibration therefore does not so much improve the model as select its operating point, trading forecast amplitude for grid accuracy at the cost of detection recall in the sparsest counties, where aggressively scaled outputs fall below the cluster-extraction threshold. This is the precision--recall knob of Section~\ref{sec:experiment}: the calibrated configuration serves intensity products, and relaxing the calibration recovers recall for detection-focused deployment. In sum, the exposure anchor and the calibration bracket the design. The anchor is the shared foundation of accuracy and localization, and the calibration is the tunable trade between them. The mixture, attention, SE, and EMA components each add incremental localization fidelity at little or no error cost. Figures~\ref{fig:baseline_comparison}(h) and~\ref{fig:ablation_bars}(c) make this balance visual. Against the learned baselines, \ours{} attains both the lowest error and the highest F\(_1\); against its own ablations it occupies the knee of the trade-off front, where no single design change improves one objective without degrading the other.

\section{Conclusion}\label{sec:conclusion}
We introduced \ours{}, a unified deep learning framework that folds emerging-hotspot detection, short-term forecasting of intensity and explicit centroids, and interpretable life-cycle tracking into a single model. Its multi-channel grid representation fuses severity-stratified crash intensity, hotspot cluster structure, temporal cyclicity, and long-run exposure. A compact CNN--Transformer backbone, refined by a regime-routed mixture-of-experts and read out by a structured intensity head, anticipates nascent high-risk regions while preserving their geometry. The head anchors a background risk on the county exposure map and adds a Hawkes-style self-excitation term under a zero-inflated negative-binomial likelihood; metric-aligned checkpoint selection and a per-county calibration then align it with the reported measures, and a minimal-cost matching scheme turns each forecast into a five-phase life-cycle trajectory. Trained once and evaluated statewide across six heterogeneous Wisconsin counties, a single model achieves the best grid-level forecasting accuracy among five identically trained baselines and leads on hotspot localization as well, on both centroid displacement and F\(_1\), with the best per-county F\(_1\) in both dense-urban counties. Just as important, it acts proactively, flagging emerging high-risk cells before they consolidate and rendering their evolution as stable, interpretable phase sequences rather than opaque snapshots.

Several limitations remain. Forecast and tracking fidelity fall in very sparse regimes, where weak signals make birth detection and phase transitions less reliable, and marginal clusters can oscillate between growth and decline without added temporal smoothing. Complex interactions such as cluster merges and splits are handled only indirectly, which can fragment trajectories and blur identity continuity. Centroid drift, most severe when adjacent hotspots compete, propagates into downstream phase labels. Reliance on historical patterns also limits responsiveness to abrupt regime shifts absent an adaptation mechanism, and the matching thresholds, smoothing parameters, and loss weights must be tuned to trade sensitivity against stability. The hybrid modeling and matching pipeline is fast enough for near-real-time analytics, yet ultra-low-latency deployment would need further optimization.

Future work will address these limitations along several lines. Adaptive, signal-aware weighting and uncertainty quantification can modulate confidence across dense and sparse contexts. Explicit modeling of merges and splits, through soft associations or graph-based continuity refinements, should improve trajectory coherence. Exogenous signals such as weather, special events, and traffic volume, together with richer crash-data modalities such as narrative reports~\cite{ma2026agentic} and infrastructure-centric scene video~\cite{gan2026crashsight}, can sharpen detection and forecasting under nonstationary conditions. Online updating and domain adaptation would let the system track evolving patterns without full retraining. Coupling the framework with high-fidelity digital-twin simulation and connected-vehicle active-safety systems could turn calibrated forecasts into concrete, closed-loop interventions~\cite{zhang2025virtual, wu2025digital, ran2026enterprise}. Finally, uncertainty visualization, active threshold calibration, and broader geographic validation will help move the framework from an analytical prototype toward a scalable, interpretable decision-support tool, and closer to the proactive, anticipatory posture that Vision Zero demands.

\section*{Acknowledgments}
The Community Maps Predictive Analytics Tool and WisTransPortal MV4000 Crash Database were developed through sponsorship and collaboration with the Wisconsin Department of Transportation.

\printcredits

\bibliographystyle{model1-num-names}

\bibliography{refs}

\end{document}